\documentclass[sigconf]{acmart}

\copyrightyear{2024}
\acmYear{2024}
\setcopyright{acmlicensed}
\acmConference[KDD '24] {Proceedings of the 30th ACM SIGKDD Conference on Knowledge Discovery and Data Mining }{August 25--29, 2024}{Barcelona, Spain.}
\acmBooktitle{Proceedings of the 30th ACM SIGKDD Conference on Knowledge Discovery and Data Mining (KDD '24), August 25--29, 2024, Barcelona, Spain}
\acmISBN{979-8-4007-0490-1/24/08}
\acmDOI{10.1145/3637528.3671665}

\usepackage{subfigure}
\usepackage{xspace}
\usepackage{multirow}
\usepackage{xcolor}
\usepackage{amsmath}
\usepackage{amsfonts}
\usepackage{bm}
\usepackage{soul}
\usepackage[ruled]{algorithm2e}
\usepackage{balance}
\usepackage{dsfont}
\usepackage[normalem]{ulem} 

\newcommand{\rev}[1]{{\color{black}{#1}}}

\newcommand{\eat}[1]{}

\newcommand{\zhuzhou}{{\sc Zhuzhou}\xspace}
\newcommand{\baoding}{{\sc Baoding}\xspace}
\newcommand{\model}{{\sc ASeer}\xspace}

\newcommand{\eg}{\emph{e.g.},\xspace}
\newcommand{\ie}{\emph{i.e.},\xspace}
\newcommand{\wrt}{\emph{w.r.t.}\xspace}
\newcommand{\aka}{\emph{a.k.a.}\xspace}
\newcommand{\etc}{\emph{etc.}\xspace}

\newcommand\figref[1]{Figure~\ref{#1}}

\newcommand\tabref[1]{Table~\ref{#1}}
\newcommand\secref[1]{Section~\ref{#1}}
\newcommand\equref[1]{Eq.~(\ref{#1})}

\newcommand{\tabincell}[2]{\begin{tabular}{@{}#1@{}}#2\end{tabular}}  

\begin{document}

\title{Irregular Traffic Time Series Forecasting Based on Asynchronous Spatio-Temporal Graph Convolutional Networks}

\author{Weijia Zhang}
\affiliation{%
  \institution{HKUST(GZ)}
  \city{Guangzhou}
  \country{China}
}
\email{vegazhang3@gmail.com}

\author{Le Zhang}
\affiliation{%
  \institution{Baidu Research}
  \city{Beijing}
  \country{China}
}
\email{zhangle0202@gmail.com}

\author{Jindong Han}
\affiliation{%
  \institution{HKUST}
  \city{Hong Kong}
  \country{China}
}
\email{jhanao@connect.ust.hk}

\author{Hao Liu}
\authornote{Corresponding author.}
\affiliation{%
  \institution{HKUST(GZ) \& HKUST}
  \city{Guangzhou}
  \country{China}
}
\email{liuh@ust.hk}

\author{Yanjie Fu}
\affiliation{%
  \institution{Arizona State University}
  \city{Phoenix}
  \country{United States}
}
\email{yanjie.fu@asu.edu}

\author{Jingbo Zhou}
\affiliation{%
  \institution{Baidu Research}
  \city{Beijing}
  \country{China}
}
\email{zhoujingbo@baidu.com}

\author{Yu Mei}
\affiliation{%
  \institution{Baidu Inc.}
  \city{Beijing}
  \country{China}
}
\email{whqyqy@hotmail.com}

\author{Hui Xiong}
\authornotemark[1]
\affiliation{%
  \institution{HKUST(GZ) \& HKUST}
  \city{Guangzhou}
  \country{China}
}
\email{xionghui@ust.hk}

\renewcommand{\shortauthors}{Weijia Zhang et al.}

\begin{abstract}
Accurate traffic forecasting is crucial for the development of Intelligent Transportation Systems~(ITS), playing a pivotal role in modern urban traffic management. Traditional forecasting methods, however, struggle with the irregular traffic time series resulting from adaptive traffic signal controls, presenting challenges in asynchronous spatial dependency, irregular temporal dependency, and predicting variable-length sequences.
To this end, we propose an \textbf{A}synchronous \textbf{S}patio-t\textbf{E}mporal graph convolutional n\textbf{E}two\textbf{R}k~(\model) tailored for irregular traffic time series forecasting.
Specifically, we first propose an Asynchronous Graph Diffusion Network to capture the spatial dependency between asynchronously measured traffic states regulated by adaptive traffic signals. 
After that, to capture the temporal dependency within irregular traffic state sequences, a personalized time encoding is devised to embed the continuous time signals. Then, we propose a Transformable Time-aware Convolution Network, which adapts meta-filters for time-aware convolution on the sequences with inconsistent temporal flow.
Additionally, a Semi-Autoregressive Prediction Network, comprising a state evolution unit and a semi-autoregressive predictor, is designed to predict variable-length traffic sequences effectively and efficiently.
Extensive experiments on a newly established benchmark demonstrate the superiority of \model compared with twelve competitive baselines across six metrics. 
\footnote{This project is available at \url{https://github.com/usail-hkust/ASeer}.}
\end{abstract}


  
\maketitle

\section{Introduction}\label{sec:intro}
\begin{figure}[tb]
  \centering
  \vspace{1mm}
  \includegraphics[width=0.98\columnwidth]{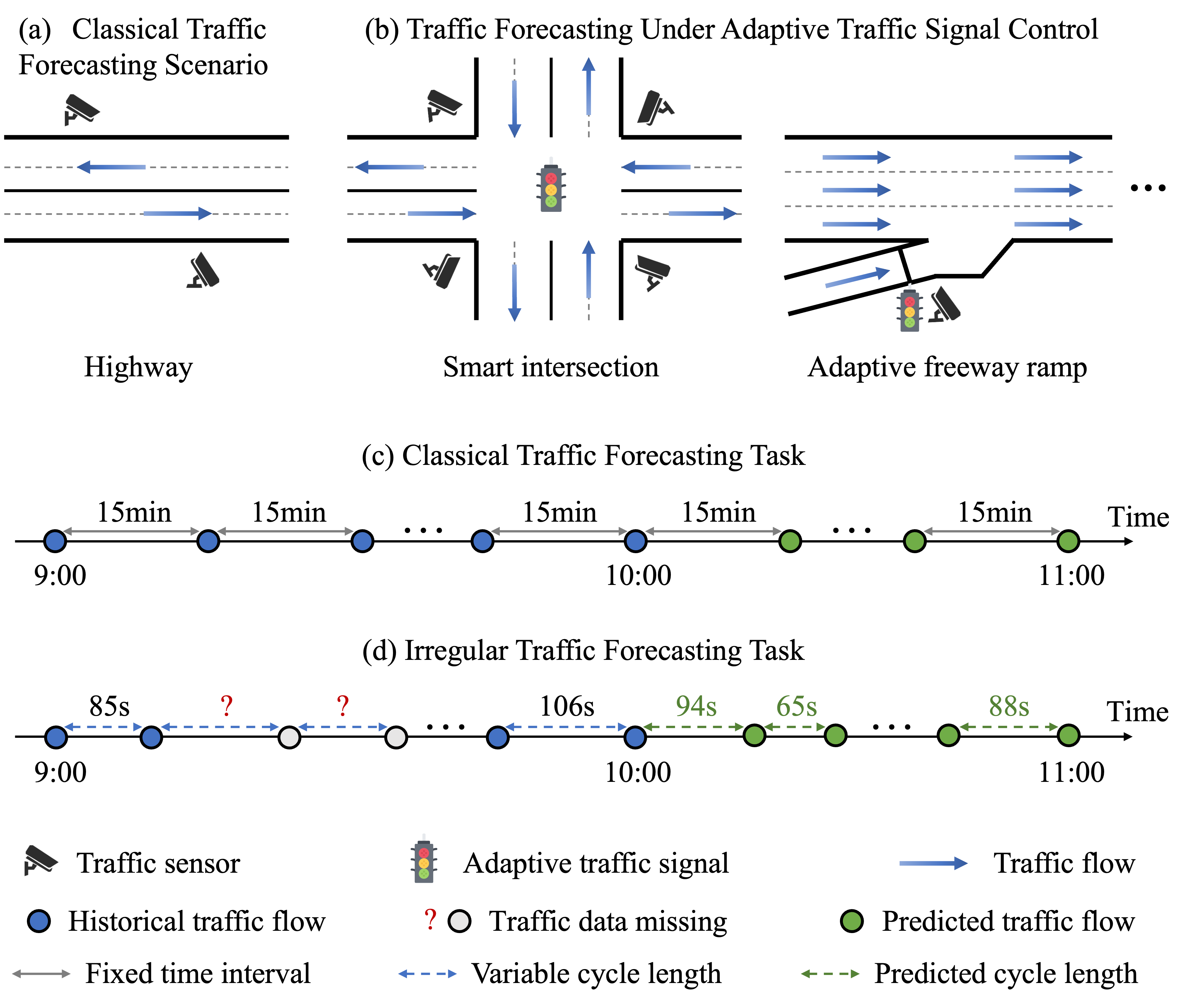}
  \vspace{-3mm}
  \caption{The distinction between classical traffic forecasting on the highway and irregular traffic forecasting under adaptive traffic signal control. The forecasting task aims to predict future traffic variations~(10:00-11:00) based on past observations~(9:00-10:00).  
  }
  \vspace{-5mm}
  \label{fig:problem}
\end{figure}

Recent years have witnessed significant advancements in traffic forecasting, which plays a pivotal role in underpinning Intelligent Transportation Systems~(ITS)~\cite{lana2018road}, facilitating emergency response and management~\cite{rahman2023deep}, and is integral to the development of autonomous driving~\cite{kohli2016traffic}.
In particular, timely and accurate traffic forecasting is of great importance to help the Intelligent Traffic Signal Control Systems (ITSCS) to anticipate future traffic state variations, thereby providing crucial insights to support the systematic analysis, informed decisions, and optimal control optimization of ITSCS to enhance the overall transportation system efficiency~\cite{wang2018review}.

In practice, the traffic dynamics of the road network is jointly decided by the vehicles on the road and the intervention of traffic signals, \eg~intersection traffic lights, ramp metering lights, and lane allocation signals~\cite{wei2019survey,sun2023hierarchical,wang2016dynamic,chen2019adaptive}.
On the one hand, the traffic signal adaptively adjusts its control cycles in response to real-time traffic flow variations~\cite{wei2019survey}.
On the other hand, traffic flows are dynamically regulated by these adaptive signal control strategies with varying cycle lengths. 
As a result, the urban traffic states, entangling both length-varying traffic signal cycles and the corresponding traffic flows, exhibit significant irregularity and render more complex traffic dynamics, as depicted in \figref{fig:problem}.

However, existing studies~\cite{jiang2022graph,li2018diffusion,yu2018spatio,wu2019graph} on urban traffic forecasting primarily focus on capturing spatiotemporal dependencies among geo-distributed time series with evenly spaced temporal variables, largely overlooking the irregularity of traffic time series induced by the interplay between traffic dynamics and adaptive traffic control policies.
\rev{These approaches model spatiotemporal dependencies and forecast traffic variations within fixed time intervals, which are misaligned with the traffic control cycles, \eg a fixed 5-minute window may always fail to cover multiple complete control cycles, leading to substantial fluctuations and inaccuracies in understanding traffic dynamics and forecasting outcomes.}
In this study, we investigate the irregular traffic time series forecasting~(\aka irregular traffic forecasting) task, aiming to predict a variable-length sequence of traffic states, encompassing traffic signal cycle lengths and
the corresponding traffic flows, in the future time window based
on incomplete historical traffic state sequences characterized by
variable time intervals.

It is a non-trivial task due to the following three major challenges:
(1)~\emph{Asynchrony in spatial dependency modeling.}
Traffic time series has obvious spatial dependency due to the traffic state's diffusion nature on road network~\cite{li2018diffusion}. \rev{However, due to the time-misaligned traffic signal cycles (diverged cycle beginning time and length) between sensors, their traffic state measurements under the adaptive control policy would be observed asynchronously. 
Such asynchrony hinders correlating and integrating these sensors' traffic states synchronously favored by classical traffic forecasting methods~\cite{li2018diffusion,yu2018spatio}, presenting a substantial challenge to model their spatial dependency.}
(2)~\emph{Irregularity in temporal dependency modeling.}
The future traffic states are correlated with their historical values. 
Unlike previous traffic forecasting research~\cite{li2018diffusion,yu2018spatio} that deals with regular traffic time series, we need to handle irregular traffic state sequences characterized by variable time intervals between successive measurements. These irregularities stem from fluctuating cycle lengths and data missing in sensor measurements, \rev{leading to inconsistent temporal flow within the time series}, challenging classical traffic forecasting methods to capture the underlying temporal dynamics and dependencies precisely.
(3)~\emph{Variable-length sequence to be predicted.}
Our goal is to predict the complete traffic state sequences in a future time window~(\eg the next hour).
However, due to signal cycle lengths varying across different sensors and times, the lengths of sequences to be predicted also vary. While an autoregressive prediction model may seem plausible for predicting sequences of variable lengths, significant error accumulation and poor prediction efficiency issues are pronounced with longer sequences~\cite{moreno2023deep, mariscalearning}. This presents a significant challenge in effectively and efficiently forecasting variable-length traffic state sequences.

To tackle the above challenges, we present an {\textbf{A}synchronous \textbf{S}patio-t\textbf{E}mporal graph convolutional n\textbf{E}two\textbf{R}k}~(\model) for irregular traffic forecasting.
Specifically, by linking traffic sensors via a traffic diffusion graph, 
we propose an Asynchronous Graph Diffusion Network to model the spatial dependencies among nodes with time-misaligned traffic state measurements. 
It allows each node to asynchronously diffuse its traffic measurements to neighbors and store received traffic information with a message buffer. The stored messages are then integrated through an asynchronous graph convolution for the spatial node representation. 
To capture temporal dependencies in irregular sequences, a learnable personalized time encoding is first devised to embed the continuous time of traffic measurements. Then, we propose a Transformable Time-aware Convolution Network that learns meta-filters to derive time-aware convolution filters with transformable filter sizes, which are applied for efficient temporal convolution on \rev{sequences with inconsistent temporal flow}. 
Lastly, we design a Semi-Autoregressive Prediction Network to iteratively predict variable-length traffic state sequences effectively and efficiently. It incorporates a state evolution unit to evolve traffic hidden state with elapsed time and a semi-autoregressive predictor to predict a sequence of consecutive traffic states at each prediction step.

Our major contributions can be summarized as follows: 
(1)~We investigate a novel irregular traffic forecasting problem, which imposes three critical new challenges for traffic forecasting from spatial, temporal, and predicted sequence length perspectives. 
(2)~We propose an Asynchronous Graph Diffusion Network to model spatial dependency among asynchronous time series data.
(3)~We propose a Transformable Time-aware Convolution Network with personalized time encoding to efficiently capture temporal dependency within irregular time series.
(4)~We design a Semi-Autoregressive Prediction Network to empower effective and efficient prediction for variable-length time series.
(5)~We meticulously collect and develop two novel real-world datasets of irregular traffic time series from two leading pilot cities for ITSCS in China, and establish a systematic evaluation scheme comprising six metrics, setting a new benchmark in the field and potentially fostering advancements in related areas. Extensive experiments demonstrate the superiority of \model compared with twelve competitive baseline approaches.

\section{Preliminaries}\label{sec:preliminary}
Consider a set of $N$ traffic sensors, denoted as $\mathbf{V} = \{v^1, v^2, \dots, v^{N}\}$, positioned on lanes governed by adaptive traffic signals, \eg~lanes connecting to smart intersections. Each sensor gathers real-time traffic data specific to a lane.

\noindent \emph{\textbf{Definition 1: Traffic State Measurement.}}
\emph{
The $n$-th chronological traffic state measurement of a sensor $v^i$ is defined as \rev{$x_n^i = \langle p_n^i, f_n^i \rangle$}, where $p^i_n$ denotes the traffic signal cycle length under the adaptive control, and $f^i_n$ is the traffic flow during this signal cycle. 
We further define $b^i_n$ and $t^i_n$ as the beginning and end timestamps~(in second) of this signal cycle, and we have $t^i_n = b^i_n + p_n^i - 1$.
}

As traffic signal cycles occur consecutively in real-world scenarios, we have $b^i_{n+1} = t^i_{n} + 1$ in case of no missing traffic states between $x_n^i$ and $x_{n+1}^i$.
Due to the unpredictable systematic failures of sensors, there could be multiple traffic states missing between two successive observed measurements. 


\noindent \emph{\textbf{Problem: Irregular Traffic Forecasting.}}
\emph{Given a historical time window $\mathcal{T}$, \eg one hour, before current timestamp $t$, and a set of historical traffic state measurements \rev{$\mathbf{X}_{[t-\mathcal{T}+1:t]}=\{[x_n^i]_{n=1}^{T^i}\}_{i=1}^N$} of all sensors $\mathbf{V}$ obtained during $\mathcal{T}$, 
our problem is to predict the complete traffic states \rev{$\mathbf{Y}_{[t+1:t + \mathcal{\tau}]}=\{[x_n^i]_{n=T^i+1}^{T^i+L^i}\}_{i=1}^N$} for all sensors in the next $\tau$ time window, e.g., the next hour, formalized as:
\begin{equation} 
    \mathcal{F}(\mathbf{X}_{[t-\mathcal{T}+1:t]}) \stackrel{}{\longrightarrow} \mathbf{Y}_{[t+1:t+\tau]},
\end{equation}
\rev{where $T^i$ is the number of observed historical measurements of $v^i$, $L^i$ is the number of ground truth traffic states of $v^i$ during the predicted time window}, $\mathcal{F}(\cdot)$ represents the forecasting model we aim to learn. 
}
\rev{Note that we use the subscript $[a:b]$ to indicate a time window spanning from timestamp $a$ to $b$.}

\begin{figure*}[tb]
  \centering
  \vspace{-2mm}
  \includegraphics[width=2\columnwidth]{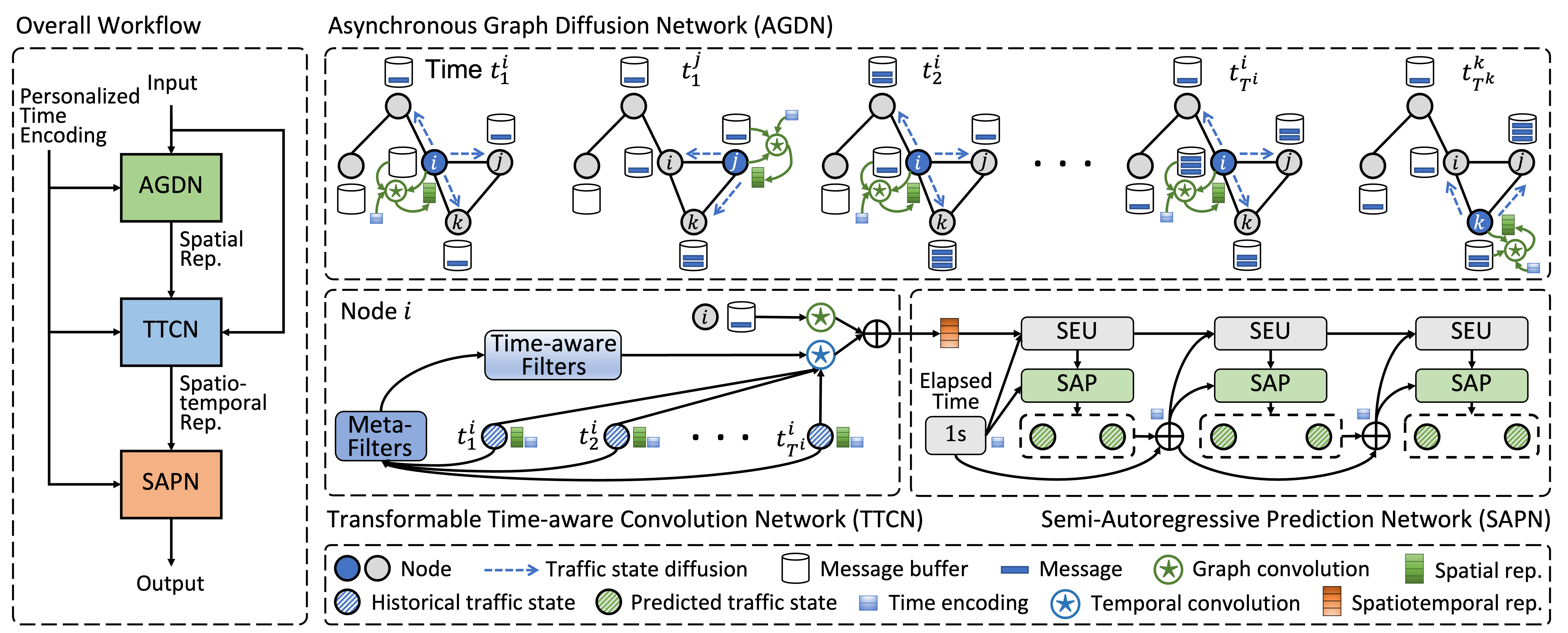}
  \vspace{-2mm}
  \caption{The framework overview of \model, which consists of three major components: Asynchronous Graph Diffusion Network (AGDN), Transformable Time-aware Convolution Network (TTCN), and Semi-Autoregressive Prediction Network (SAPN). The traffic states are first inputted to AGDN to obtain spatial representations, which are incorporated by TTCN to acquire the spatiotemporal representations. After that, SAPN predicts the variable-length traffic state sequence based on the spatiotemporal representations. Throughout the entire process, personalized time encoding is used to embed continuous time.}
  \vspace{-2mm}
  \label{fig:model}
\end{figure*}

\section{Methodology}
\noindent \textbf{Framework Overview.}
\figref{fig:model} shows the framework overview of \model, which consists of three major components.
Specifically, {Asynchronous Graph Diffusion Network}~(AGDN) models asynchronous spatial dependency based on a traffic diffusion graph. 
When a node~(\ie~traffic sensor) has a traffic state measurement, AGDN asynchronously diffuses the node's traffic measurement to its neighbors, which receive and store the diffused traffic state into their message buffers. Next, the node performs an asynchronous graph convolution to obtain spatial representation through attentively integrating the stored traffic messages, and then the buffer will be cleared. 
After that, a {Transformable Time-aware Convolution Network}~(TTCN) is adopted to model the temporal dependency within irregular traffic state sequences. TTCN learns meta-filters to derive time-aware convolution filters with transformable filter sizes based on spatial representations obtained from AGDN and traffic measurements along with personalized time encoding. Then the derived time-aware convolution filters are applied for efficient temporal convolution on irregular traffic state sequences to acquire the spatiotemporal representation for each node. 
Finally, a {Semi-Autoregressive Prediction Network}~(SAPN) is devised to iteratively predict variable-length traffic state sequences. In each prediction step, a {State Evolution Unit}~(SEU), whose hidden state is initialized by spatiotemporal representations, is first introduced to evolve each node’s future traffic hidden state with the elapsed time, then a {Semi-Autoregressive Predictor}~(SAP) is adopted to predict a sequence of consecutive traffic states based on both evolutionary and initial traffic hidden states, as well as predicted elapsed time.


\subsection{Asynchronous Spatial Dependency Modeling}
Previous traffic forecasting studies model spatial dependency by introducing graph neural networks to synchronously diffuse and aggregate time-aligned traffic states between different sensor nodes~\cite{li2018diffusion, yu2018spatio}. 
However, in our problem, the observed traffic state measurements of different sensors cannot be aligned due to the distinct timestamps of their traffic signal cycles and the data missing issue, which causes severe asynchrony in spatial dependency modeling.

To this end, by linking sensors via a traffic diffusion
graph, we propose an {Asynchronous Graph Diffusion Network}~(AGDN), \rev{as illustrated in \figref{fig:model}}, to model asynchronous spatial dependency between time-misaligned traffic measurements. 
We detail it below.

\noindent \textbf{Diffusion Graph Construction.}
To model spatial dependency between traffic sensors, we construct a traffic diffusion graph $\mathcal{G}=(\mathcal{V},\mathcal{X}_{\mathcal{V}},\mathcal{E},\mathcal{X}_{\mathcal{E}})$, where the graph nodes $\mathcal{V}=\mathbf{V}$ represents a set of sensors, $\mathcal{X}_{\mathcal{V}} = \mathbf{X}_{[t-\mathcal{T}+1:t]}$ denotes features of nodes $\mathcal{V}$, $\mathcal{E}$ are a set of edges indicating proximity between nodes, and $\mathcal{X}_{\mathcal{E}}$ are features in edges $\mathcal{E}$. 
Specifically, we define proximity $e^{ij} \in \mathcal{E}$ as: 
$e^{ij}=1$ if the geographical distance between node $v^i$ and $v^j$ is smaller than a threshold $\epsilon$, otherwise $e^{ij}=0$,
and there is no self-loop for each node. 
We also define some edge features $x_e^{ij} \in \mathcal{X}_\mathcal{E}$ between nodes $v^i$ and $v^j$, including geographical distance and the direct reachability in the lane-level road network.
Note that it is not limited to geographical proximity and reachability, other graph construction approaches can also be embraced.

\noindent \textbf{Asynchronous Diffusion and Storage.}
Assume a traffic state measurement $x^j_{n^-}$ of node $v^j$ is observed at timestamp $t^j_{n^-}$, then $v^j$ will diffuse $x^j_{n^-}$ as a traffic message to its adjacent nodes $v^i \in \mathcal{N}_j$ in terms of edges $\mathcal{E}$, which can be more formally denoted as:
\begin{equation} 
\text{AsynDiff}\Big(v^j \stackrel{x^j_{n^-}}{\longrightarrow} \{v^i: \forall v^i \in \mathcal{N}_j\}\Big).
\end{equation}
For each node $v^i \in \mathcal{N}_j$, it receives the traffic message $x^j_{n^-}$ and then stores it into its message buffer $\mathcal{B}^i$ for later use:
\begin{equation} 
\text{Store}\Big(x^j_{n^-} \stackrel{}{\longrightarrow} \{\mathcal{B}^i: \forall v^i \in \mathcal{N}_j\}\Big).
\end{equation}
Since the timestamps of traffic state measurements are misaligned for different nodes, the traffic messages' diffusion and storage processes perform in an asynchronous way.

\noindent \textbf{Asynchronous Graph Convolution.}
An immediate problem is how to exploit traffic messages stored in the message buffer to enhance each node's spatial perception. We achieve this by enforcing each node $v^i$ asynchronously integrates the traffic messages in its message buffer $\mathcal{B}^i$ via an asynchronous graph convolution operation, which is triggered when a measurement $x^i_n$ is observed.

Specifically, we first employ $x^i_n$ to query the message buffer $\mathcal{B}^i$ for the proximity weights computation with each traffic message $x^j_{{n^-}} \in \mathcal{B}^i$ via the following attention operation:
\begin{equation} 
\begin{gathered}
\alpha_{nn^{-}}=\frac{\text{exp}(\beta_{nn^{-}})}{\sum_{x^j_{n^\prime} \in \mathcal{B}^i} \text{exp}(\beta_{nn^{\prime}})}, \\
\beta_{nn^{-}}=\mathbf{v}^\top \tanh \left(\mathbf{W}_a \big[x_{n}^i \oplus x_{n^{-}}^j \oplus \phi^i(t^i_n - t^j_{n^{-}}) \oplus x_e^{ij}\big]\right),
\end{gathered}
\label{equ:attention}
\end{equation}
where $\oplus$ indicates concatenation operation, $\mathbf{v}$ and $\mathbf{W}_a$ are learnable parameters, and $\phi^i(\cdot)$ is a learnable time encoding function to embed cycle-related patterns for each node that will be detailed in the next section.

Once the proximity weights are obtained, we asynchronously integrate node's stored traffic messages received from neighbors via an attentive graph convolution to obtain the spatial representation:
\begin{equation}  
\widetilde{h}^i_{n} = \text{MLP}\left(\sum_{x^j_{n^-} \in \mathcal{B}^i} \alpha_{nn^-} \cdot \left[x^j_{n^-} \oplus \phi^i(t^i_n - t^j_{n^{-}}) \oplus x_e^{ij} \right]\right),
\label{equ:aggregation}
\end{equation}
where MLP represents a multi-layer perceptron. It is noteworthy that after each asynchronous graph convolution operation on $\mathcal{B}^i$, all the traffic messages in it will be cleared. It indicates that each node only integrates adjacent traffic messages from its last traffic measurement's timestamp to the current measurement's timestamp $t^i_n$, which guarantees each message is utilized exactly once to avoid redundant information and computation.

There could be some messages received and stored in the message buffer $\mathcal{B}^i$ after timestamp $t^i_{T^i}$ of the last observed traffic state measurement of node $v^i$ during historical time window $\mathcal{T}$.
Thus, we perform a similar asynchronous graph convolution operation for these remaining messages by adding a virtual measurement $\widebar{x}^i_{T^i}$ at timestamp $t^i_{T^i}$ without traffic state values. The obtained spatial representation is denoted as $\widebar{h}^i_{T^i}$.

\subsection{Irregular Temporal Dependency Modeling}
Convolutional Neural Network~(CNN)~\cite{lecun2015deep} is widely applied to classical traffic forecasting tasks for its both efficiency and effectiveness in temporal dependency modeling~\cite{yu2018spatio,guo2019attention,wu2019graph,lan2022dstagnn}. 
However, applying CNN to our task faces two problems.
First, CNN fails to directly process irregular traffic sequences with variable sequence lengths. 
Second, CNN is incompetent to model temporal dependency in the sequence with varying time intervals, \rev{as its filter parameters are fixed and cannot adaptively adjust according to the inconsistent temporal flow of sequence, which leads to distinct patterns of sequence dependencies.}


To tackle the above problems, we propose a {Transformable Time-aware Convolution Network}~(TTCN) \rev{that enables to model irregular sequences with transformable time-aware convolution filters},
and further devise a personalized time encoding function to embed the unique cycle-related patterns for each node. 
Specifically, given the historical time window $\mathcal{T}$ before $t$, for each node $v^i$,
we first concatenate the traffic state measurement $x^i_{n}$ during $\mathcal{T}$ with the corresponding spatial representation $\widetilde{h}^i_{n}$ and the encoding of time intervals $\phi^i(t^i_{T^i}-t^i_n)$ from timestamp $t^i_n$ to timestamp $t^i_{T^i}$ of the last observed traffic measurement:
\begin{equation} 
    z^i_{n}=\left[x^i_{n} \oplus \widetilde{h}^i_{n} \oplus \phi^i(t^i_{T^i}-t^i_n)\right].
\end{equation}

\subsubsection{\textbf{Personalized Time Encoding}}
The desired time encoding should not only indicate the absolute time interval but also imply the unique cycle-related patterns of traffic dynamics in different nodes. 
For example, a time interval may signify a distinct number of traffic signal cycles for different sensors, which is important for temporal dependency modeling, especially when the time interval spans multiple missing traffic states.
Inspired by the positional encoding in Transformer~\cite{vaswani2017attention}, we introduce a personalized time encoding by adopting a learnable trigonometric function to embed the time interval $\Delta t$ for each node:
\begin{equation} 
\phi_p^i(\Delta t)[s]=\left\{
\begin{array}{lr}
\Delta t,\quad\quad\quad\quad\ \text{if}\ \ s=0\\
\sin(\omega^i_k \Delta t),\quad  \text{if}\ \ s=2k+1\\
\cos(\omega^i_k \Delta t),\quad  \text{if}\ \ s=2k+2
\end{array}
\right.,
\label{equ:pte}
\end{equation}
where the above equation denotes $s$-th element of the time encoding $\phi_p^i(\Delta t) \in \mathbb{R}^{d_\phi + 1}$, and $\omega^i_{k}$ are learnable parameters, indicating the cyclical characteristics of this function. Each node has an individual time encoding function with separate parameters so that can learn its unique cycle-related patterns.

Due to the data missing problem, some nodes may have too sparse measurement data to learn a satisfactory unique time encoding function. 
Hence, we also jointly learn a generic time encoding $\phi_g(\Delta t)$, which has a similar function expression to \equref{equ:pte} but is shared by all nodes. 
Then, we introduce a learnable weight $\lambda_i$ for each node to adaptively integrate the above two time encoding:
\begin{equation} 
\phi^i(\Delta t)= (1-\exp(-\lambda_i^2))\cdot\phi_p^i(\Delta t) + \exp(-\lambda_i^2)\cdot\phi_g(\Delta t).
\end{equation}
$\lambda_i$ is initialized to be close to zero so that the nodes with limited or even no available data can weigh more on generic time encoding.

\subsubsection{\textbf{Transformable Time-aware Convolution Network}}
This section assumes all the operations performed on node $v^i$, thus we omit the superscript $i$ to ease the presentation.
We first define \rev{$\mathbf{z}_{[t-\mathcal{T}+1:t]}=[z_{1}, \cdots, z_{T}]$} and $T$ as the sequence length.
\rev{As illustrated in \figref{fig:model},} we leverage meta-filters to derive the time-aware convolution filters with dynamic parameters and transformable filter size $T$ based on sequence inputs, formulated as:
\begin{equation} 
\begin{gathered}
    \mathbf{f}_d=\left[\text{Norm}\left(\mathbf{F}_d(z_{1})\right), \cdots, \text{Norm}\left(\mathbf{F}_d(z_{T})\right)\right],\\
    \rev{\text{Norm}\left(\mathbf{F}_d(z_{n})\right) = \frac{\text{exp}(\mathbf{F}_d(z_{n}))}{\sum_{z_{n^{\prime}} \in \mathbf{z}_{[t-\mathcal{T}+1:t]}} \text{exp}(\mathbf{F}_d(z_{n^{\prime}}))},}
\end{gathered}
\label{equ:metafilter}
\end{equation}
where $\mathbf{f}_d \in \mathbb{R}^{T \times D_{in}}$ is the derived filter for $d$-th feature map, and $\mathbf{F}_d$ denotes a meta-filter that can be instantiated by learnable neural networks. We normalize the derived filter parameters along the temporal dimension to ensure consistent scaling of the convolution results for variable-length sequences.

With $D$ filters derived according to \equref{equ:metafilter}, we obtain the traffic sequence representation $h_{T}\in \mathbb{R}^D$ via the following temporal convolution operation:
\begin{equation} 
\begin{gathered}
    h_{T} = \left[\mathbf{z}_{[t-\mathcal{T}+1:t]} \star \mathbf{f}_1, \cdots, \mathbf{z}_{[t-\mathcal{T}+1:t]} \star \mathbf{f}_D \right],\\
    \mathbf{z}_{[t-\mathcal{T}+1:t]} \star \mathbf{f}_d = \sum_{n=1}^{T} \mathbf{f}_d[n]^\top \mathbf{z}_{[t-\mathcal{T}+1:t]}[n],
\end{gathered}
\label{equ:convolution}
\end{equation}
where $\star$ denotes the convolution operation. 
Then we attain the overall spatiotemporal representation for each node via the representations integration: \rev{$\mathbf{h}_T = h_{T} + \widebar{h}_{T}$}.

\rev{Compared to CNN, TTCN has several advantages in adaptively modeling sequences with inconsistent temporal flow.}
First, the derived filter is transformable according to sequence length, which enables it to adaptively process variable-length sequences. 
\rev{Moreover, it can derive tailored parameterized filters for sequences with changeable temporal flow or other characteristics.}
Furthermore, it is worth noting that as the learnable parameters of meta-filter are independent of sequence length, TTCN is allowed to directly model the long-term temporal dependency via an arbitrarily large-size convolution filter without increasing any filter parameters.

\subsection{Variable-Length Traffic Sequence Prediction}
Our goal is to predict the complete traffic state sequences, including a sequence of traffic signal cycle lengths and the corresponding traffic flows, for all nodes in a future time window. 
However, the sequences to be predicted have variable lengths in terms of the differences in sensors, time windows, or prediction algorithms, and the sequence lengths cannot be known in advance. 
While an autoregressive prediction model that iteratively predicts the next step's value based on previously predicted values seems feasible for the variable-length sequence prediction, the prediction for long sequence can lead to severe error accumulation and poor prediction efficiency issues in this approach~\cite{moreno2023deep, mariscalearning}.

\rev{To tackle the above problems, as displayed in \figref{fig:model}, we design a {Semi-Autoregressive Prediction Network}~(SAPN) to iteratively predict sub-sequences until the complete sequence meets the requirements of the task. It not only enables variable-length sequence prediction in an efficient way but also mitigates the error accumulation issue for long sequence prediction.}
\rev{Since prediction processes are the same for all nodes, we omit the superscript $i$ to ease the presentation as well.}

To be specific, we employ the spatiotemporal representation \rev{$\mathbf{h}_T$} acquired from AGDN and TTCN as the initial traffic hidden state.
In each prediction step, a semi-autoregressive predictor predicts a sequence of consecutive traffic states based on the evolutionary and initial traffic hidden states, as well as the predicted elapsed time encoding, formulated as:
\begin{equation} 
\left[\hat{p}_n, \hat{u}_n \right]_{n=T+m\xi+1}^{T+(m+1)\xi} = \text{SAP}\left(\big[\hat{\mathbf{h}}_{T+m\xi+1} \oplus \mathbf{h}_{T} \oplus \phi(\hat{\delta}_{T+m\xi+1}) \big]\right),
\end{equation}
where $\xi$ is the prediction step size, $m\geq0$ denotes $m$-th prediction step, \rev{$\left[\hat{p}_n, \hat{u}_n \right]_{n=T+m\xi+1}^{T+(m+1)\xi}$ respectively represent a sequence of consecutive cycle lengths and unit time~(per second) traffic flows}, and $\hat{\delta}_{T+m\xi+1}$ indicates the elapsed time to the timestamp $t_T$ of node's last observed measurement. $\hat{\delta}_{T+m\xi+1}$ is initialized to $1$ when $m=0$, and iteratively updates based on the accumulation of predicted cycle lengths:
\vspace{-2mm}
\begin{equation} 
\hat{\delta}_{T+(m+1)\xi+1} = \hat{\delta}_{T+m\xi+1} + \sum_{k=1}^\xi{\hat{p}_{{T+m\xi+k}}}.
\vspace{-1mm}
\end{equation}
Since the underlying traffic state is actually dynamically evolving with passage of time, we introduce a state evolution unit that learns to evolve each node’s traffic hidden state with the elapsed time:
\vspace{-1mm}
\begin{equation} 
\hat{\mathbf{h}}_{{T+m\xi+1}} = \text{SEU}\left(\hat{\mathbf{h}}_{{T+(m-1)\xi+1}}, \phi(\hat{\sigma}_{T+m\xi+1})\right), 
\vspace{-1mm}
\end{equation}
where \rev{$\hat{\mathbf{h}}_{{T+(m-1)\xi+1}}=\mathbf{h}_T$} and $\hat{\sigma}_{T+m\xi+1}=1$ if $m=0$, otherwise $\hat{\sigma}_{T+m\xi+1}=\sum_{k=1}^\xi{\hat{p}_{{T+(m-1)\xi+k}}}$, \rev{representing the elapsed time to last update of traffic hidden state.}
Next, we can obtain the corresponding traffic flows of predicted traffic signal cycles by multiplying the predicted unit time traffic flows with cycle lengths:
\vspace{-1mm}
\begin{equation} 
\left[\hat{f}_n\right]_{n=T+m\xi+1}^{T+(m+1)\xi} = \left[\hat{u}_n\right]_{n=T+m\xi+1}^{T+(m+1)\xi} \odot \left[\hat{p}_n\right]_{n=T+m\xi+1}^{T+(m+1)\xi},
\end{equation}
where $\odot$ denotes Hadamard product.
By iteratively performing the above prediction step until the predicted sequence covers the required time window, we can derive the variable-length traffic state sequence we expect.

Compared to autoregressive models, our SAPN predicts a variable-length sequence with fewer prediction steps, which improves prediction efficiency and may reduce the risks of causing prediction error accumulation. 
It is evident that both the autoregressive and non-autoregressive prediction models can be regarded as a special case of semi-autoregressive model when the prediction step size is set to one or the length of sequence. Thus, our SAPN can also be considered as incorporating both strengths of autoregressive and non-autoregressive models to predict variable-length sequences. 
\rev{In the implementation, we instantiate SAP via MLP and SEU via Gated Recurrent Unit~\cite{chung2014empirical} as its mechanism aligns well with the recurrent update for traffic hidden state.}

\subsection{Model Training}
Due to the data missing problem, we design three masked losses to train our model. 
The first loss is introduced to optimize the traffic signal cycle length forecasting via the masked Mean Absolute Error~(MAE):
\vspace{-2mm}
\begin{equation} 
\mathcal{L}_p = \frac{1}{\sum_{i=1}^N L^i_\mathds{1} }\sum_{i=1}^N\sum_{l=1}^{L^i} \left|\hat{p}^i_{{T^i+l}} - p^i_{{T^i+l}}\right| \times \zeta^i_{{T^i+l}},
\vspace{-1mm}
\end{equation}
$\zeta^i_{{T^i+l}}$ is a mask term, which equals zero if the ground truth value $p^i_{{T^i+l}}$ is missing, otherwise it equals one, and $L^i_\mathds{1}$ denotes the number of nonzero mask items for each node.

To further mitigate error accumulation in cycle length prediction, we additionally introduce a timing loss to improve the accuracy of predicted elapsed time accumulated by cycle lengths:
\vspace{-1mm}
\begin{equation} 
\mathcal{L}_{\delta} = \frac{1}{\sum_{i=1}^N L^i_\mathds{1}} \sum_{i=1}^N\sum_{l=1}^{L^i} \left|\hat{\delta}^i_{{T^i+l}} - \delta^i_{{T^i+l}} \right| \times \zeta^i_{{T^i+l}}.
\vspace{-1mm}
\end{equation}

Similarly, we introduce a masked MAE loss to optimize the corresponding traffic flow prediction for each traffic signal cycle:
\vspace{-1mm}
\begin{equation} 
\mathcal{L}_f = \frac{1}{\sum_{i=1}^N L^i_\mathds{1}}\sum_{i=1}^N\sum_{l=1}^{L^i} \left|\hat{u}^i_{{T^i+l}} \times p^i_{{T^i+l}} - f^i_{{T^i+l}} \right| \times \zeta^i_{{T^i+l}}.
\vspace{-1mm}
\end{equation}
Since traffic flow prediction is also based on cycle lengths, to avoid disturbance from the error of predicted cycle lengths, we use the ground truth cycle lengths to calculate the corresponding traffic flows in the training phase.

Consequently, \model aims to jointly minimize an overall objective that combines the above three masked losses:
\begin{equation} 
\mathcal{L} = \mathcal{L}_p + \mathcal{L}_{\delta} + \mathcal{L}_f.
\label{equ:hybridloss}
\end{equation}

\section{Experiments}\label{sec:exp}
\subsection{\textbf{Experimental Setup}}
\noindent \textbf{Datasets.}
We conduct experiments on two real-world datasets, \zhuzhou and \baoding, which represent two major pilot cities for ITSCS and autonomous driving in China. 
Both datasets consist of a set of entrance lanes connecting to smart intersections and traffic state measurements of lanes collected by the installed camera sensors.
The statistics of the datasets are summarized
in \tabref{table:dataset}.
We take the data from the first $60\%$ of the entire time range as the training set, the following $20\%$ for validation, and the remaining $20\%$ as the test set. 
For both datasets, we set both historical and predicted time window lengths $\mathcal{T}$ and $\tau$ to one hour.
Please refer to \secref{app:dataset} for more description and analysis of the datasets.

\noindent \textbf{Implementation Details.}
All experiments are performed on a Linux server with 20-core Intel(R) Xeon(R) Gold 6148 CPU @ 2.40GHz and NVIDIA Tesla V100 GPU.
We calculate spherical distance as the geographical distance and choose distance threshold $\epsilon=1$km and prediction step size $\xi=12$. 
\rev{We adopt three layers MLPs for asynchronous graph convolution}, semi-autoregressive predictor, and meta-filters. 
The dimension for time encoding is set to $d_{\phi}=16$, and dimensions for convolution filters, state evolution unit, and hidden layers of the above MLPs are all set to $64$.
To reduce parameter magnitude, in the implementation, we individualize the last layer's parameters but share the other parameters of MLP for meta-filters.
We employ Adam optimizer to train our model, set learning rate to $0.001$. \model and all learnable baselines are trained with an early stop criterion if the loss doesn't decrease lower on the validation set over $10$ epochs.

\begin{table}[tb]
    \small
    \centering
    \caption{Statistics of datasets.}
    \vspace{-4mm}
    \setlength{\tabcolsep}{1mm}{
        \begin{tabular}{l|c|c}
            \toprule[1pt]
            \textbf{Description} & \zhuzhou & \baoding\\
                \hline
                \# of measurements & 19,824,504 & 13,093,975\\
                \hline
                \# of sensors & 620 & 264\\
                \hline
            Time range & 2022/07/20-2022/10/02 & 2021/12/01-2022/02/25 \\
                \hline
            Missing period ratio & 44.2\% & 27.2\% \\
                \hline
                \multirow{3}{*}{\tabincell{l}{Average / maximal\\ ground truth sequence\\ length to be predicted\\}} & \multirow{3}{*}{57 / 213} & \multirow{3}{*}{64 / 155} \\
                & & \\
                & & \\
            \bottomrule[1pt]
    \end{tabular}}
    \vspace{-4mm}
    \label{table:dataset}
\end{table}

\begin{table*}[tb]
\small
\centering
\vspace{-2mm}
\caption{Overall performance evaluated by C-MAE, C-RMSE, C-MAPE, F-MAE, F-RMSE, and F-AAE on \zhuzhou and \baoding. The best-performing results are highlighted in \textbf{bold}, and the second-best results are highlighted by \underline{underline}.}
\vspace{-4mm}
\setlength{\tabcolsep}{2mm}{\begin{tabular}{l|cccccc|cccccc}
	\toprule[1pt]
	\multirow{2}{*}{\textbf{Algorithm}} & \multicolumn{6}{c|}{\zhuzhou} &
	\multicolumn{6}{c}{\baoding} \\ \cline{2-13}
	& C-MAE & C-RMSE & C-MAPE & F-MAE & F-RMSE & F-AAE & C-MAE & C-RMSE & C-MAPE & F-MAE & F-RMSE & F-AAE\\ 
        \hline
        LAST &  50.5386 & 135.2616 & 5.54\% & 1.6669 & 3.0995 & 0.9192 & 42.8037 & 106.6547 & 4.79\% & 1.7521 & 2.8031  & 0.9557 \\
        HA &  52.1532 & 135.3569 & 5.76\% & 1.4502 & 2.6567  & 0.7998 & 49.7496 & 114.8265 & 5.53\% & 1.5449 & 2.4594  & 0.8427 \\
        TCN &  43.7838 & 110.1670 & 5.01\% & 1.3950 & 2.5824 & 0.7818 & 35.8318 & 95.7333 & 4.18\% & 1.3815 & 2.2060 & 0.7635 \\
        GRU &  40.6209 & 99.8693 & 4.82\% & 1.3623 & 2.5553 & 0.7524 & 30.4621 & 83.4349 & 3.82\% & 1.3576 & 2.1655 & 0.7423 \\
        T-LSTM &  39.1882 & 87.3458 & 5.38\% & 1.3641 & 2.5494 & 0.7539 & 29.0845 & 82.5219 & 3.76\% & 1.3673 & 2.1887 & 0.7475 \\
        GRU-D &  37.8531 & 84.6255 & 5.23\% & 1.3486 & 2.5333 & 0.7449 & 28.9117 & 82.5226 & 3.67\% & 1.3611 & 2.1735 & 0.7456 \\
        mTAND &  37.5762 & 86.3045 & \textbf{3.93\%} & 1.3563 & 2.5282 & 0.7498 & 27.2703 & 78.1066 & \underline{2.86\%} & 1.3575 & 2.1641 & 0.7487 \\
        Warpformer & \underline{35.7369} & 85.3125 & 5.02\% & 1.3399 & 2.5662 & 0.7405 & 27.8527 & 78.7168 & 3.89\% & 1.3554 & 2.1720 & 0.7427 \\
        DCRNN &  38.5976 & 90.3190 & 4.36\% & 1.3318 & \underline{2.4438} & 0.7348 & 31.0564 & 76.3693 & 3.86\% & 1.3681 & 2.1601 & 0.7467 \\
        GWNet &  38.9913 & 106.6415 & 4.52\% & 1.3834 & 2.7915 & 0.7618 & \underline{26.4988} & 84.3211 & 3.05\% & 1.3925 & 2.2482 & 0.7903 \\
        STAEformer & 40.4448 & \underline{78.2176} & 4.84\% & 1.3503 & 2.4501 & 0.7468 & 28.1453 & 73.5397 & 3.74\% & 1.3801 & 2.1518 & 0.7585 \\
        PDFormer & 36.4779 & 86.1173 & 4.66\% & \underline{1.3170} & 2.4575 & \underline{0.7269} & 27.1969 & \underline{68.5613} & 3.82\% & \underline{1.3540} & \underline{2.1496} & \underline{0.7413} \\
        \hline
        \model &  \textbf{32.5803} & \textbf{72.1835} & \underline{4.10\%} & \textbf{1.2913} & \textbf{2.3864} & \textbf{0.7151} & \textbf{19.1188} & \textbf{54.4919} & \textbf{2.80\%} & \textbf{1.3062} & \textbf{2.0827} & \textbf{0.7219} \\
	\bottomrule[1pt]
\end{tabular}}
\vspace{-2mm}
\label{table:overall_performance}
\end{table*}

\noindent \textbf{Evaluation Metrics.}
We define six metrics to comprehensively evaluate the forecasting performance, including C-MAE, C-RMSE, and C-MAPE to evaluate the accuracy of predicted traffic signal cycle lengths, and F-MAE, F-RMSE, and F-AAE for the traffic flow prediction evaluation. Lower is better for all these metrics.

For the evaluation of signal cycle length prediction, we quantify the predicted errors of both the beginning timestamps and cycle lengths via masked Mean Absolute Error~(MAE), Root Mean Squared Error~(RMSE), and Mean Absolute Percentage Error~(MAPE):
\vspace{-3mm}
\begin{equation} 
\begin{gathered}
\text{C-MAE} = \frac{1}{2 \sum_{i=1}^N K^i_\mathds{1}} \sum_{i=1}^N\sum_{k=1}^{K^i} \left(|\hat{b}^i_{{k}} - b^i_{{k}}| + |\hat{p}^i_{{k}} - p^i_{{k}}| \right) \times \zeta^i_{{k}},\\
\text{C-RMSE} = \sqrt{\frac{1}{2 \sum_{i=1}^N K^i_\mathds{1}} \sum_{i=1}^N\sum_{k=1}^{K^i} \left( (\hat{b}^i_{{k}} - b^i_{{k}})^2 + (\hat{p}^i_{{k}} - p^i_{{k}})^2 \right) \times \zeta^i_{{k}}},\\
\text{C-MAPE} = \frac{100\%}{2 \sum_{i=1}^N K^i_\mathds{1}} \sum_{i=1}^N\sum_{k=1}^{K^i} \left(\frac{|\hat{b}^i_{k} - b^i_{{k}}|}{{\delta}^i_k} + \frac{|\hat{p}^i_{{k}} - p^i_{{k}}|}{p^i_{{k}}} \right) \times \zeta^i_{{k}}.
\end{gathered}
\end{equation}
where $K^i$ denotes the number of ground truth traffic states of each sensor for evaluation, $K^i_\mathds{1}$ is the number of observed measurements.

Since traffic flows depend on the corresponding traffic signal cycles, it is incomparable between the predicted and ground truth traffic flows if they are misaligned in signal cycles. 
Thus, we introduce two types of metrics to evaluate the prediction accuracy of traffic flows from multiple aspects. 
First, we assume all the traffic signal cycle lengths can be accurately predicted and use the ground truth cycle lengths for calculation, then we can directly compare the predicted and ground truth traffic flows via the following masked MAE and RMSE metrics:
\vspace{-1mm}
\begin{equation} 
\begin{gathered}
\text{F-MAE} = \frac{1}{\sum_{i=1}^N K^i_\mathds{1}} \sum_{i=1}^N\sum_{k=1}^{K^i} \left|\hat{f}^i_{_{k}} - f^i_{{k}}\right| \times \zeta^i_{{k}},\\
\text{F-RMSE} = \sqrt{\frac{1}{\sum_{i=1}^N K^i_\mathds{1}} \sum_{i=1}^N\sum_{k=1}^{K^i}  \left(\hat{f}^i_{_{k}} - f^i_{{k}} \right)^2  \times \zeta^i_{{k}}}.\\
\end{gathered}
\end{equation}
Second, without the above assumption for cycle lengths, by computing traffic flow density at any timestamp, we calculate the masked Accumulative Absolute Error~(AAE) between predicted and ground truth traffic flow density at identical timestamps:
\vspace{-1mm}
\begin{equation} 
\text{F-AAE} = \frac{1}{\mathbf{Z}} \sum_{i=1}^N \sum_t \left|\hat{\rho}^i_t - {\rho}^i_{t}\right| \times \eta^i_{t},
\end{equation}
where $\hat{\rho}^i_t = \hat{f}^i_{_k} / \hat{p}^i_{_k}, t\in[\hat{b}^i_k, \hat{t}^i_k]$ and $\rho^i_t = f^i_{k} / p^i_{k}, t\in[b^i_k, t^i_k]$
are the predicted and ground truth traffic flow densities of sensor $v^i$ at timestamp $t$, respectively. $\eta^i_{t}$ is the mask term at $t$, which equals one if $\rho^i_t$ can be obtained from observed measurement, and zero otherwise. In our experiments, the timestamp is in seconds, and we use a normalization term $\mathbf{Z}$ to obtain the average result in minutes.

\noindent \textbf{Baselines.}
We compare our approach with the following twelve baselines, including two heuristic approaches~(LAST, HA), two classical sequence modeling approaches~(TCN~\cite{bai2018empirical}, GRU~\cite{chung2014empirical}), four competitive irregular time series modeling approaches~(T-LSTM~\cite{baytas2017patient}, GRU-D~\cite{che2018recurrent}, mTAND~\cite{shukla2020multi}, Warpformer~\cite{warpformer2023}), and four competitive classical traffic forecasting approaches~(DCRNN~\cite{li2018diffusion}, GWNet~\cite{wu2019graph}, STAEformer~\cite{liu2023spatio}, PDFormer~\cite{pdformer2023}).
\rev{For fair comparison, all learnable baseline models are set to predict the cycle lengths and unit time traffic flows by optimizing the hybrid loss function in \equref{equ:hybridloss} like \model.} 
In addition, except for autoregressive models~(GRU, T-LSTM, GRU-D, DCRNN), other baselines predict in a semi-autoregressive manner with the same prediction step size as \model. 
We carefully tune major hyper-parameters of each baseline based on their recommended settings for better performance on our datasets. Please refer to \secref{app:baselines} for more details of baselines.

\subsection{Overall Performance}
\tabref{table:overall_performance} reports the overall performance of \model and all compared baselines on two datasets \wrt six metrics. As can be seen, \model achieves the best overall performance among all the compared approaches on two datasets, which demonstrates our model's superiority in irregular traffic forecasting task.
Besides, we have several observations. 
Firstly, all learnable approaches outperform the statistical approaches~(\ie LAST, HA), which validates that the data-driven approaches to learn complex non-linear interactions within traffic data is helpful for this task.
Secondly, we find CNN-based baselines TCN and GWNet do not achieve a desired performance for the reason that classical CNN with the fixed parameterized filters is incompetent to model the temporal dependency in irregular sequences. 
Thirdly, we observe \model obtains a superior overall performance than approaches~(\ie GRU-D, T-LSTM, mTAND, and Warpformer) for irregular time series, as these approaches fail to model the complex spatial dependencies between large-scale sensors. From these approaches, we notice mTAND has a slight advantage in C-MAPE than \model on \zhuzhou. This is probably because mTAND as a powerful approach for interpolation task performs well in the short-term future cycles' beginning times prediction. However, \model significantly outperforms mTAND in the other metrics.
Lastly, we observe a notable performance improvement by comparing \model with the state-of-the-art approaches~(\ie DCRNN, GWNet, STAEformer, and PDFormer) for classical traffic forecasting. The improvement can primarily be attributed to the capability of \model to effectively model asynchronous spatial dependency and irregular temporal dependency in the irregular traffic forecasting problem.

\subsection{Ablation Study}
\begin{figure}[tb]
  \centering
  \includegraphics[width=0.95\columnwidth]{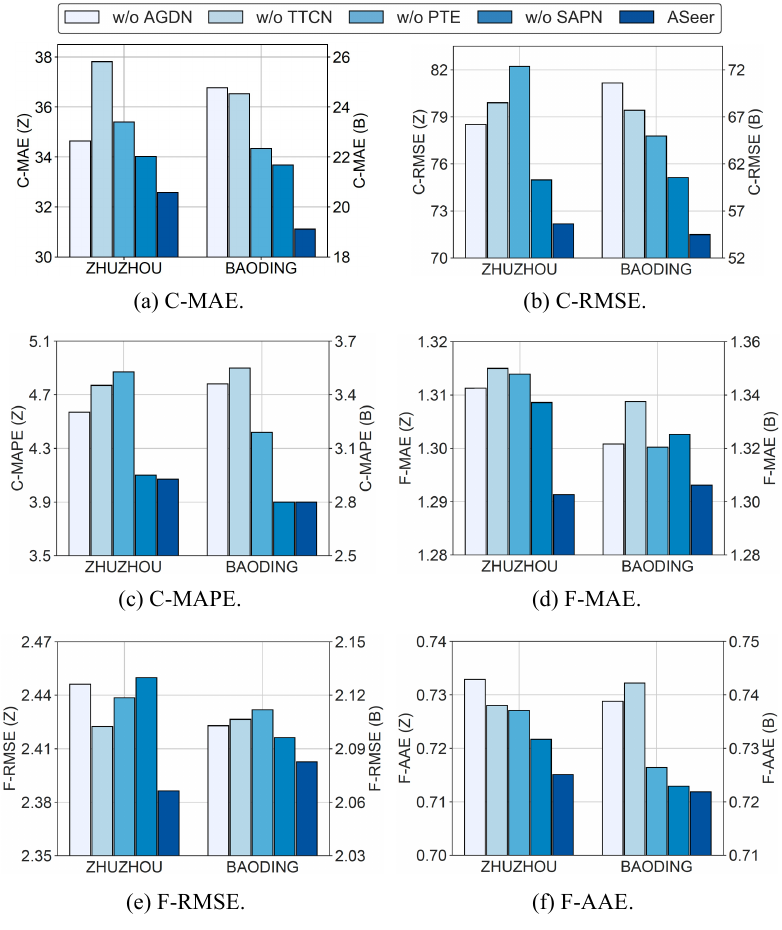}
  \vspace{-4mm}
  \caption{Results of ablation study. "Z" and "B" denote \zhuzhou and \baoding, respectively.} 
  \vspace{-3mm}
  \label{fig:ablation_study}
\end{figure}

We evaluate the performance of \model and its four variants on both \zhuzhou and \baoding across six metrics. 
(1)~\textbf{w/o AGDN} removes the AGDN module; (2)~\textbf{w/o TTCN} replaces TTCN with a 1D CNN, whose filter size is set to the maximal sequence length in the dataset; (3)~\textbf{w/o PTE} removes personalized time encoding; (4)~\textbf{w/o SAPN} replaces SAPN with an autoregressive MLP predictor. 
The results of ablation study are shown in \figref{fig:ablation_study}, As can be seen, removing any component causes notable overall performance degradation compared to \model, which demonstrates the effectiveness of each component.
From these results, we observe \textbf{w/o TTCN} almost results in significant performance descent for all metrics on both datasets, which verifies the effectiveness of TTCN to improve classical CNN to model the temporal dependency within irregular traffic sequences.
In addition, \textbf{w/o AGDN} causes a remarkable accuracy decline for all the metrics \wrt traffic flow, which validates the effect of AGDN on modeling asynchronous spatial dependency of traffic dynamics. We also observe \textbf{w/o AGDN} causes a more obvious accuracy decline on \baoding than \zhuzhou for three metrics \wrt cycle lengths. This is probably because the distribution of cycle lengths in \baoding is denser, AGDN's smoothness induces a more precise prediction. 
Moreover, we notice that \textbf{w/o PTE} leads to a consistent performance reduction for all metrics on both datasets, which demonstrates that a well-learned personalized time encoding function to embed continuous time for each sensor can facilitate the prediction of both cycle lengths and traffic flows.
Finally, by comparing \model with \textbf{w/o SAPN}, we observe a more obvious performance degradation on \baoding for metrics \wrt cycle lengths, which is probably because the sequence is longer on \baoding, an autoregressive model causes a severe error accumulation problem on cycle length prediction. \textbf{w/o SAPN} also shows a consistent performance descent for three metrics \wrt traffic flow, which confirms that SAPN improves the long traffic sequence prediction performance.

\subsection{Parameter Sensitivity}

We conduct experiments for two important hyper-parameters, \ie the prediction step size $\xi$ and dimension of all hidden layers, on both \zhuzhou and \baoding to study the sensitivity of these hyper-parameters. We report experimental results on metrics C-MAE, F-MAE, and F-AAE to evaluate the model's prediction performance on both cycle lengths and traffic flows.

\begin{figure}[tb]
\vspace{-1.5mm}
  \centering
  \subfigure[{C-MAE.}]{
    \includegraphics[width=0.305\columnwidth]{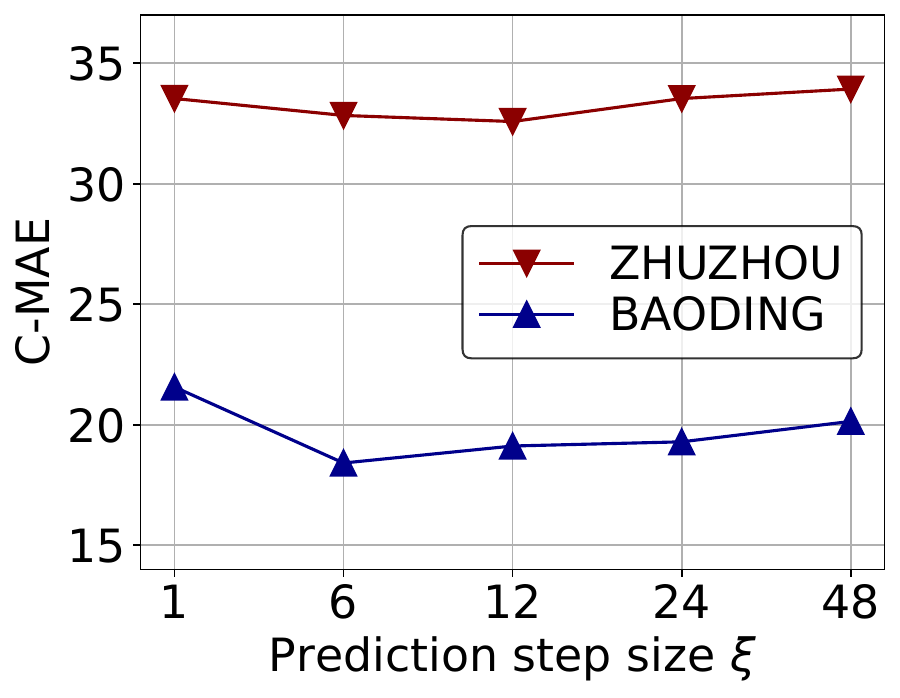}}\hspace{-1mm}
  \subfigure[{F-MAE.}]{
    \includegraphics[width=0.32\columnwidth]{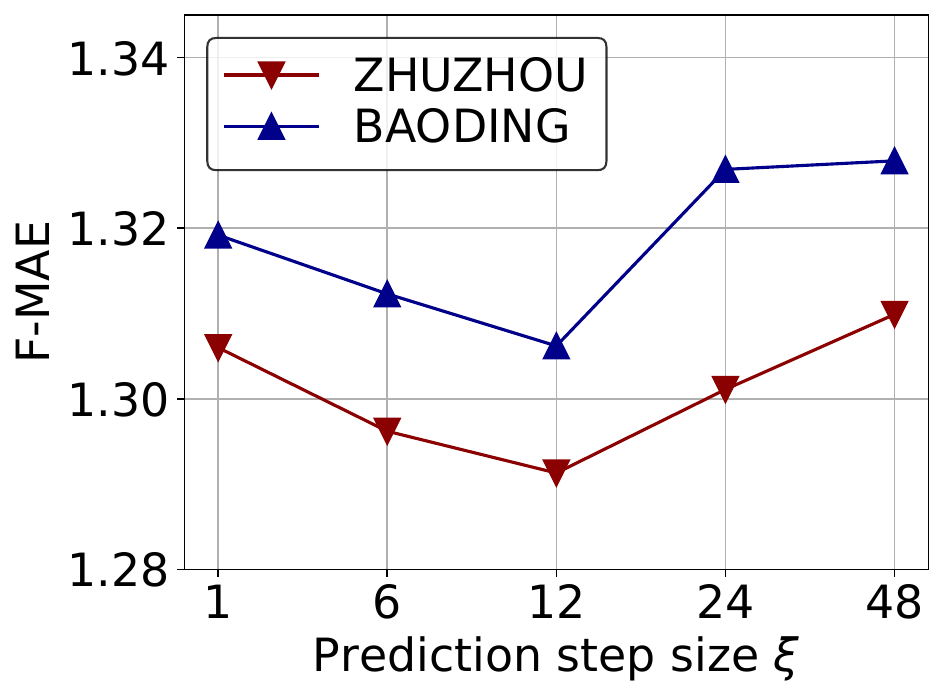}}\hspace{-0.5mm}
  \subfigure[{F-AAE.}]{
    \includegraphics[width=0.33\columnwidth]{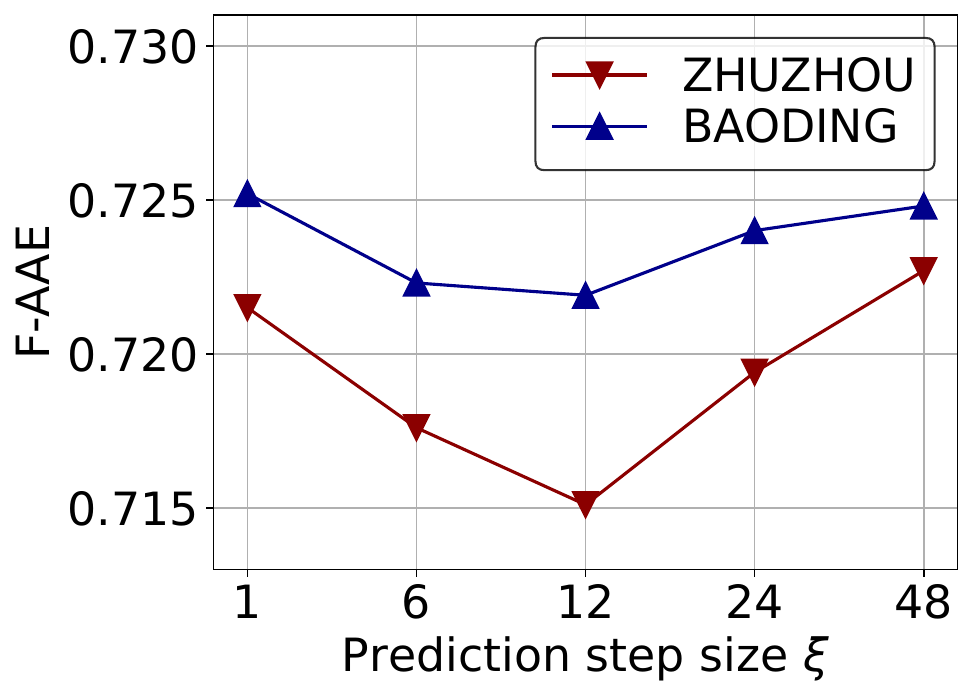}}
  \vspace{-4mm}
  \caption{Effect of different prediction step sizes.} 
  \vspace{-3mm}
  \label{fig:param_stepsize}
\end{figure}

\figref{fig:param_stepsize} shows the results of varying the prediction step size $\xi$ from $1$ to $48$. As can be seen, there is a notable overall prediction performance improvement by increasing $\xi$ from $1$~(autoregressive model) to $12$~(semi-autoregressive model), which demonstrates the effectiveness of SAPN to mitigate error accumulation problem in the autoregressive prediction model. However, we also observe a performance degradation when the prediction step size is too large. This is probably because a too-large prediction step size may result in under-training for SAPN to make predictions based on different elapsed times. 

\begin{figure}[tb]
  \centering
  \subfigure[{C-MAE.}]{
    \includegraphics[width=0.31\columnwidth]{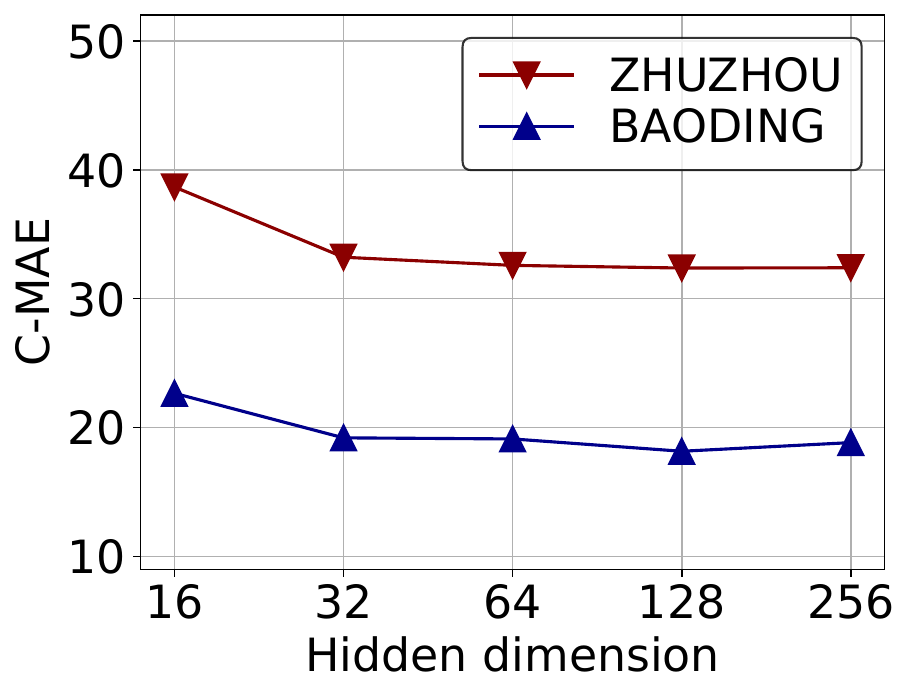}}\hspace{-1mm}
  \subfigure[{F-MAE.}]{
    \includegraphics[width=0.325\columnwidth]{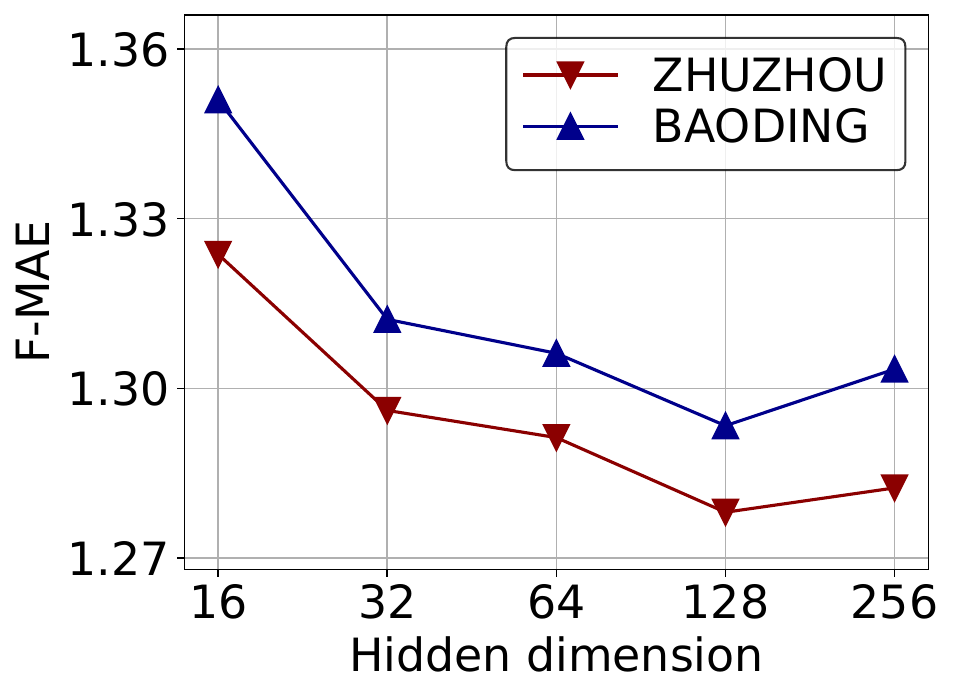}}\hspace{-0.5mm}
  \subfigure[{F-AAE.}]{
    \includegraphics[width=0.325\columnwidth]{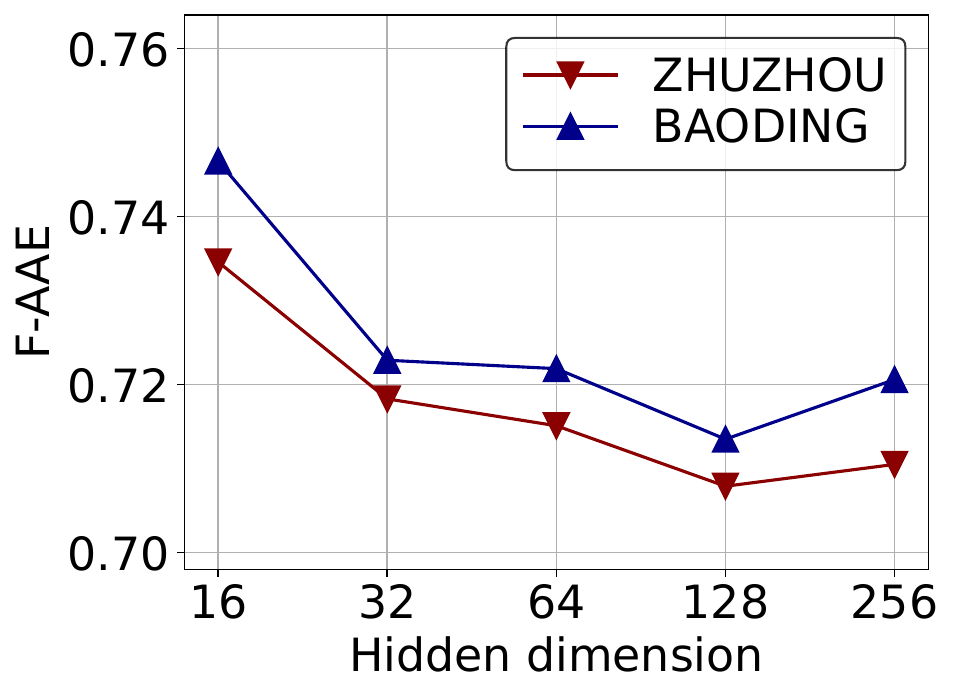}}
  \vspace{-4mm}
  \caption{Effect of different hidden dimensions.} 
  \vspace{-3mm}
  \label{fig:param_dimension}
\end{figure}

We vary the dimension of model's all hidden layers from $16$ to $256$. The results are shown in \figref{fig:param_dimension}. We can observe a remarkable prediction performance improvement by increasing the hidden dimension from $16$ to $32$, and the performance is continuously improving and achieves the best when the dimension is set to $128$. However, a larger hidden dimension also takes more expensive computational overhead. Thus, we have to balance the performance and computation cost for the selection of model's hidden dimension.

\subsection{Prediction Efficiency Analysis}
\label{sec:efficiency}
We conduct experiments to test the prediction efficiency of different models. To ensure a fair comparison, we eliminate the influence of different models on the prediction lengths by standardizing the prediction process. This involves allowing all models to predict the maximum lengths of the corresponding ground truth sequences. 

\begin{figure}[tb]
 \vspace{-0.5mm}
  \centering
  \subfigure[{\zhuzhou.}]{
    \includegraphics[width=0.465\columnwidth]{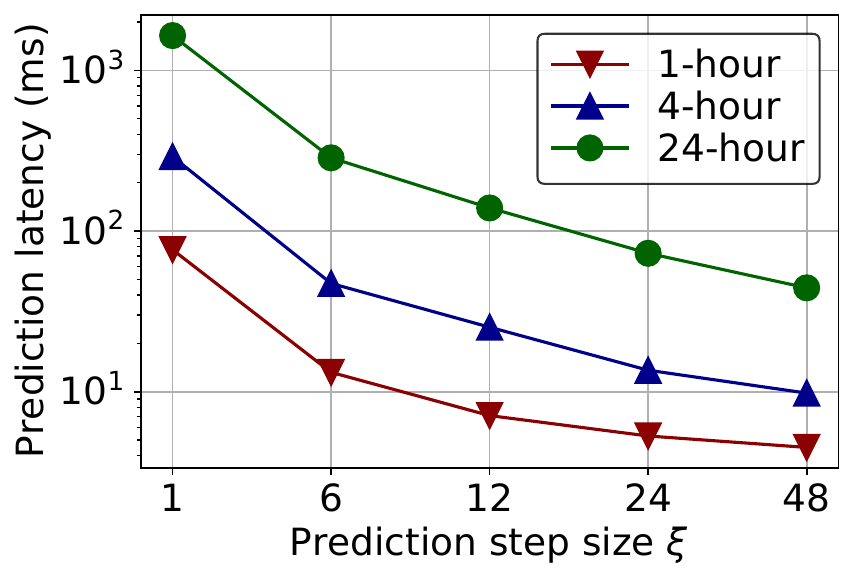}}
  \subfigure[{\baoding.}]{
    \includegraphics[width=0.465\columnwidth]{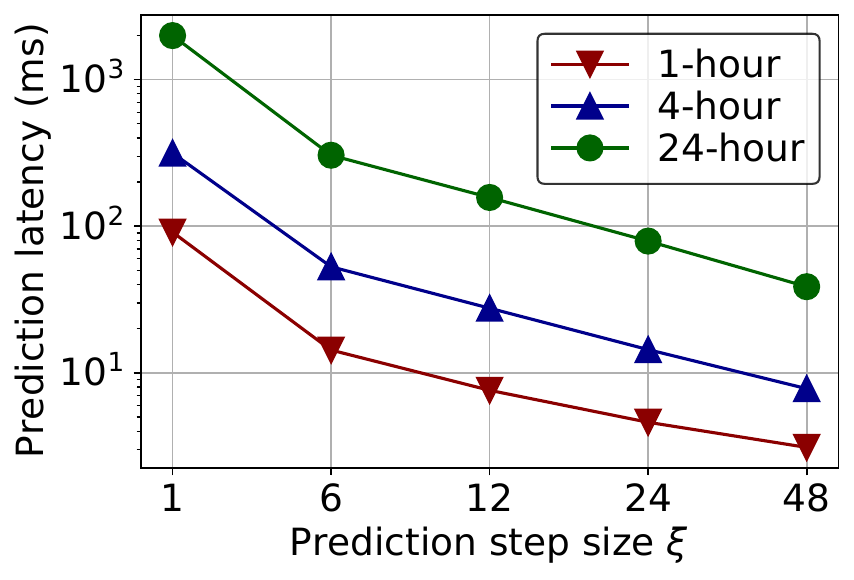}}
  \vspace{-4mm}
  \caption{Prediction latency of SAPN with different prediction step sizes.} 
  \vspace{-4mm}
  \label{fig:latency}
\end{figure}

\noindent \textbf{Efficiency of SAPN.} 
To evaluate the effect of SAPN on prediction efficiency, we conduct experiments on both \zhuzhou and \baoding to specifically test SAPN's average prediction latency based on different prediction step sizes $\xi$ from $1$ to $48$. 
We report the respective results of predicting future $1$, $4$, and $24$ hours traffic states in \figref{fig:latency}. As can be seen, the prediction latency is notably reduced by comparing semi-autoregressive models~($\xi>1$) with autoregressive model~($\xi=1$) due to the reduction of total prediction steps.
The magnitude of latency reduction even approaches the prediction step size when we predict longer sequences or the step size is not too large, which demonstrates the significant effectiveness of SAPN to improve prediction efficiency.
We also observe with the prediction step size increasing, the prediction latency is consistently reduced, and with the predicted hours rising, the model can have a significantly higher prediction efficiency by setting a larger prediction step size. This observation indicates that we can choose a larger prediction step size with the predicted sequence length increasing for higher prediction efficiency.

\begin{figure}[tb]
  \centering
    \includegraphics[width=0.8\columnwidth]{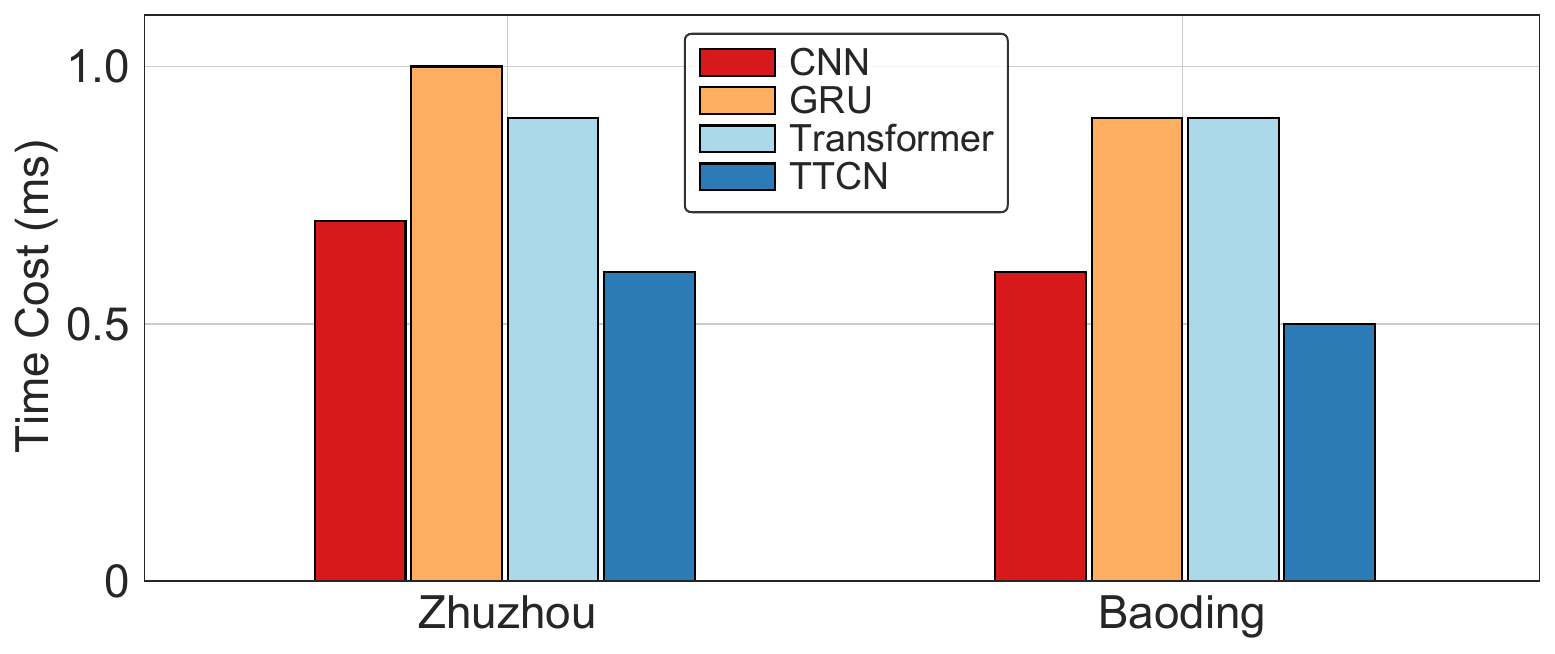}
  \vspace{-3mm}
  \caption{Efficiency of various modules in temporal modeling.} 
  \vspace{-4mm}
  \label{fig:modules}
\end{figure}

\noindent \textbf{Efficiency of TTCN.}
To study TTCN's efficiency, we replace TTCN with several commonly used modules in temporal modeling, \ie CNN, GRU, and Transformer, and test their running time costs. As illustrated in \figref{fig:modules}, TTCN achieves more than 40\% and 33\% faster results than GRU and transformer, respectively, on both datasets. Furthermore, to our surprise, TTCN exhibits even faster than CNN. This is probably because TTCN can directly handle variable-length sequences with transformable filter sizes, while CNN is limited to processing fixed-length sequences via padding or clipping, thus it may cost additional time to process longer sequences beyond their original lengths. It demonstrates the efficiency of TTCN.

\section{Related Work}
\noindent \textbf{Traffic Forecasting.}
Recently years, deep learning models have dominated the traffic forecasting tasks for their extraordinary capability in modeling the complex spatio-temporal characteristics within traffic data~\cite{zhang2017deep,yu2018spatio,guo2019attention,wu2019graph,lan2022dstagnn,yao2018deep,li2018diffusion,bai2020adaptive,zheng2020gman,fang2021spatial,shao2022decoupled,pdformer2023,zhang2020semi,han2024bigst}. For spatial modeling, a part of studies~\cite{zhang2017deep,yao2018deep} first partition a city into a grid-based region map, then utilize Convolutional Neural Networks~(CNNs) to capture spatial dependencies between adjacent regions. After that, Graph Neural Networks~(GNNs)~\cite{defferrard2016convolutional,velivckovic2018graph,wu2020comprehensive,kipf2017semi,yuan2020spatio,zhang2022multi} are widely used to model the non-euclidean spatial dependencies in traffic data~\cite{jiang2022graph,shao2022decoupled,liu2023robust, ji2023spatio,zhang2020semi,han2024bigst}. For example, studies~\cite{li2018diffusion, yu2018spatio} employ GNNs to model the traffic flow diffusion process in the road network. Studies~\cite{guo2019attention, zheng2020gman,zhang2020semi,pdformer2023,guo2021learning} incorporate attention mechanism into GNNs to learn the dynamic spatial dependencies between the road network sensors. In addition to the pre-defined relational graph derived from road networks, some works~\cite{wu2019graph,bai2020adaptive,lan2022dstagnn,yi2023fouriergnn,shao2022pre,han2024bigst} attempt to directly learn the latent graph structure from traffic data. 
For temporal modeling, CNNs~\cite{zhang2017deep,yu2018spatio,guo2019attention,wu2019graph,lan2022dstagnn} and Recurrent Neural Networks~(RNNs)~\cite{yao2018deep,li2018diffusion,bai2020adaptive} are frequently adopted to capture temporal dependencies within traffic data. Compared to RNNs, CNNs enable parallel computing for all time steps, which exhibits extreme advantages in computational efficiency.
However, all the methods of above studies are designed for the time-aligned traffic data with fixed time interval, which fails to handle the challenges of asynchronous spatial dependency and irregular temporal dependency in the irregular traffic forecasting problem.

\noindent \textbf{Irregularly Sampled Time Series.}
This work is also related to the literature about learning from irregularly sampled time series, which is a kind of time series data characterized by varying time intervals between temporally adjacent observations~\cite{zhang2021graph}.
A straightforward approach is to divide the irregularly sampled time series into a regular one with fixed time intervals~\cite{lipton2016directly}. However, such a temporal discretization method may lead to information loss and data missing problems~\cite{shukla2020multi,shukla2020survey}. 
Recent studies tend to directly learn from irregularly sampled time series. Specifically, some studies improve RNNs by using a time gate~\cite{neil2016phased}, a time decay term~\cite{che2018recurrent}, or memory decomposition mechanism~\cite{baytas2017patient} to adjust RNNs' memory update for adapting irregular time series. 
Another line of studies introduces neural Ordinary Differential Equations~(ODEs)~\cite{chen2018neural} to model the continuous dynamics in time series, and assume the latent states of time series are continuously evolving through continuous time~\cite{rubanova2019latent,schirmer2022modeling}. 
Besides, attention mechanism is also applied to model irregularly sampled time series~\cite{horn2020set, shukla2020multi, zhang2021graph, warpformer2023, wei2023compatible}. 
For example, \citet{shukla2020multi} proposes a multi-time attention network to learn embedding of continuous time.
\citet{warpformer2023} employs a doubly self-attention to learn representation from the input data unified by a warping module.
\citet{zhang2021graph} introduce GNNs to capture time-varying dependencies between sensors by performing the graph convolution operation at all timestamps when there is an observation at an arbitrary sensor. However, it will be extremely time-consuming once the data is significantly asynchronous across large-scale sensors like us. 
Furthermore, the above studies primarily focus on solving irregular time series classification instead of forecasting tasks.
Finally, to our knowledge, there are no prior studies attempting to modify CNNs to adapt to the irregular time series modeling.

\vspace{-1mm}
\section{Conclusion}\label{sec:conclusion}
In this paper, we investigated a new irregular traffic forecasting problem that aims to predict irregular traffic time series resulting from adaptive traffic signal controls, and presented \model, an Asynchronous Spatio-Temporal Graph Convolutional Network, to address this problem.
Specifically, by representing the traffic sensors as nodes and linking them via a traffic diffusion graph, we first proposed an Asynchronous Graph Diffusion Network to model the spatial dependency between the time-misaligned traffic state measurements of nodes. 
After that, to capture the temporal dependency within irregular traffic state sequences, we devised a personalized time encoding to embed the continuous time for each node and proposed a Transformable Time-aware Convolution Network to perform efficient temporal convolution on sequences with inconsistent temporal flow.
Furthermore, a Semi-Autoregressive Prediction Network was designed to iteratively predict variable-length traffic state sequences effectively and efficiently. 
Finally, extensive experiments on two newly constructed real-world datasets demonstrated the superiority of \model compared with twelve competitive baseline approaches across six metrics. 

\vspace{-1mm}
\section*{Acknowledgements}
This work was supported by the National Natural Science Foundation of China (Grant No.92370204, No.62102110), National Key R\&D Program of China~(Grant No.2023YFF0725001), Guangdong Basic and Applied Basic Research Foundation~(Grant No.2023B1515120057), Guangzhou-HKUST(GZ) Joint Funding Program~(Grant No.2023A03 J0008), Education Bureau of Guangzhou Municipality, CCF-Baidu Open Fund.

\bibliographystyle{ACM-Reference-Format}
\balance
\bibliography{sample}


\begin{thebibliography}{54}


\ifx \showCODEN    \undefined \def \showCODEN     #1{\unskip}     \fi
\ifx \showDOI      \undefined \def \showDOI       #1{#1}\fi
\ifx \showISBNx    \undefined \def \showISBNx     #1{\unskip}     \fi
\ifx \showISBNxiii \undefined \def \showISBNxiii  #1{\unskip}     \fi
\ifx \showISSN     \undefined \def \showISSN      #1{\unskip}     \fi
\ifx \showLCCN     \undefined \def \showLCCN      #1{\unskip}     \fi
\ifx \shownote     \undefined \def \shownote      #1{#1}          \fi
\ifx \showarticletitle \undefined \def \showarticletitle #1{#1}   \fi
\ifx \showURL      \undefined \def \showURL       {\relax}        \fi
\providecommand\bibfield[2]{#2}
\providecommand\bibinfo[2]{#2}
\providecommand\natexlab[1]{#1}
\providecommand\showeprint[2][]{arXiv:#2}

\bibitem[Bai et~al\mbox{.}(2020)]%
        {bai2020adaptive}
\bibfield{author}{\bibinfo{person}{Lei Bai}, \bibinfo{person}{Lina Yao},
  \bibinfo{person}{Can Li}, \bibinfo{person}{Xianzhi Wang}, {and}
  \bibinfo{person}{Can Wang}.} \bibinfo{year}{2020}\natexlab{}.
\newblock \showarticletitle{Adaptive graph convolutional recurrent network for
  traffic forecasting}.
\newblock \bibinfo{journal}{\emph{Advances in neural information processing
  systems}}  \bibinfo{volume}{33} (\bibinfo{year}{2020}),
  \bibinfo{pages}{17804--17815}.
\newblock


\bibitem[Bai et~al\mbox{.}(2018)]%
        {bai2018empirical}
\bibfield{author}{\bibinfo{person}{Shaojie Bai}, \bibinfo{person}{J~Zico
  Kolter}, {and} \bibinfo{person}{Vladlen Koltun}.}
  \bibinfo{year}{2018}\natexlab{}.
\newblock \showarticletitle{An empirical evaluation of generic convolutional
  and recurrent networks for sequence modeling}.
\newblock \bibinfo{journal}{\emph{arXiv preprint arXiv:1803.01271}}
  (\bibinfo{year}{2018}).
\newblock


\bibitem[Baytas et~al\mbox{.}(2017)]%
        {baytas2017patient}
\bibfield{author}{\bibinfo{person}{Inci~M Baytas}, \bibinfo{person}{Cao Xiao},
  \bibinfo{person}{Xi Zhang}, \bibinfo{person}{Fei Wang},
  \bibinfo{person}{Anil~K Jain}, {and} \bibinfo{person}{Jiayu Zhou}.}
  \bibinfo{year}{2017}\natexlab{}.
\newblock \showarticletitle{Patient subtyping via time-aware LSTM networks}. In
  \bibinfo{booktitle}{\emph{Proceedings of the 23rd ACM SIGKDD Conference on
  Knowledge Discovery and Data Mining}}. \bibinfo{pages}{65--74}.
\newblock


\bibitem[Che et~al\mbox{.}(2018)]%
        {che2018recurrent}
\bibfield{author}{\bibinfo{person}{Z Che}, \bibinfo{person}{S Purushotham},
  \bibinfo{person}{K Cho}, \bibinfo{person}{D Sontag}, {and} \bibinfo{person}{Y
  Liu}.} \bibinfo{year}{2018}\natexlab{}.
\newblock \showarticletitle{Recurrent Neural Networks for Multivariate Time
  Series with Missing Values.}
\newblock \bibinfo{journal}{\emph{Scientific reports}} (\bibinfo{year}{2018}),
  \bibinfo{pages}{6085--6085}.
\newblock


\bibitem[Chen et~al\mbox{.}(2019)]%
        {chen2019adaptive}
\bibfield{author}{\bibinfo{person}{Jiming Chen}, \bibinfo{person}{Weixin Lin},
  \bibinfo{person}{Zidong Yang}, \bibinfo{person}{Jianyuan Li}, {and}
  \bibinfo{person}{Peng Cheng}.} \bibinfo{year}{2019}\natexlab{}.
\newblock \showarticletitle{Adaptive ramp metering control for urban freeway
  using large-scale data}.
\newblock \bibinfo{journal}{\emph{IEEE Transactions on Vehicular Technology}}
  \bibinfo{volume}{68}, \bibinfo{number}{10} (\bibinfo{year}{2019}),
  \bibinfo{pages}{9507--9518}.
\newblock


\bibitem[Chen et~al\mbox{.}(2018)]%
        {chen2018neural}
\bibfield{author}{\bibinfo{person}{Ricky~TQ Chen}, \bibinfo{person}{Yulia
  Rubanova}, \bibinfo{person}{Jesse Bettencourt}, {and} \bibinfo{person}{David
  Duvenaud}.} \bibinfo{year}{2018}\natexlab{}.
\newblock \showarticletitle{Neural ordinary differential equations}. In
  \bibinfo{booktitle}{\emph{Proceedings of the 32nd International Conference on
  Neural Information Processing Systems}}. \bibinfo{pages}{6572--6583}.
\newblock


\bibitem[Chung et~al\mbox{.}(2014)]%
        {chung2014empirical}
\bibfield{author}{\bibinfo{person}{Junyoung Chung}, \bibinfo{person}{Caglar
  Gulcehre}, \bibinfo{person}{Kyunghyun Cho}, {and} \bibinfo{person}{Yoshua
  Bengio}.} \bibinfo{year}{2014}\natexlab{}.
\newblock \showarticletitle{Empirical evaluation of gated recurrent neural
  networks on sequence modeling}. In \bibinfo{booktitle}{\emph{NIPS Workshop on
  Deep Learning}}.
\newblock


\bibitem[Defferrard et~al\mbox{.}(2016)]%
        {defferrard2016convolutional}
\bibfield{author}{\bibinfo{person}{Micha{\"e}l Defferrard},
  \bibinfo{person}{Xavier Bresson}, {and} \bibinfo{person}{Pierre
  Vandergheynst}.} \bibinfo{year}{2016}\natexlab{}.
\newblock \showarticletitle{Convolutional neural networks on graphs with fast
  localized spectral filtering}.
\newblock \bibinfo{journal}{\emph{Advances in neural information processing
  systems}} (\bibinfo{year}{2016}), \bibinfo{pages}{3844--3852}.
\newblock


\bibitem[Fang et~al\mbox{.}(2021)]%
        {fang2021spatial}
\bibfield{author}{\bibinfo{person}{Zheng Fang}, \bibinfo{person}{Qingqing
  Long}, \bibinfo{person}{Guojie Song}, {and} \bibinfo{person}{Kunqing Xie}.}
  \bibinfo{year}{2021}\natexlab{}.
\newblock \showarticletitle{Spatial-temporal graph ode networks for traffic
  flow forecasting}. In \bibinfo{booktitle}{\emph{Proceedings of the 27th ACM
  SIGKDD Conference on Knowledge Discovery and Data Mining}}.
  \bibinfo{pages}{364--373}.
\newblock


\bibitem[Guo et~al\mbox{.}(2019)]%
        {guo2019attention}
\bibfield{author}{\bibinfo{person}{Shengnan Guo}, \bibinfo{person}{Youfang
  Lin}, \bibinfo{person}{Ning Feng}, \bibinfo{person}{Chao Song}, {and}
  \bibinfo{person}{Huaiyu Wan}.} \bibinfo{year}{2019}\natexlab{}.
\newblock \showarticletitle{Attention based spatial-temporal graph
  convolutional networks for traffic flow forecasting}. In
  \bibinfo{booktitle}{\emph{Proceedings of AAAI conference on artificial
  intelligence}}. \bibinfo{pages}{922--929}.
\newblock


\bibitem[Guo et~al\mbox{.}(2021)]%
        {guo2021learning}
\bibfield{author}{\bibinfo{person}{Shengnan Guo}, \bibinfo{person}{Youfang
  Lin}, \bibinfo{person}{Huaiyu Wan}, \bibinfo{person}{Xiucheng Li}, {and}
  \bibinfo{person}{Gao Cong}.} \bibinfo{year}{2021}\natexlab{}.
\newblock \showarticletitle{Learning dynamics and heterogeneity of
  spatial-temporal graph data for traffic forecasting}.
\newblock \bibinfo{journal}{\emph{IEEE Transactions on Knowledge and Data
  Engineering}} \bibinfo{volume}{34}, \bibinfo{number}{11}
  (\bibinfo{year}{2021}), \bibinfo{pages}{5415--5428}.
\newblock


\bibitem[Han et~al\mbox{.}(2024)]%
        {han2024bigst}
\bibfield{author}{\bibinfo{person}{Jindong Han}, \bibinfo{person}{Weijia
  Zhang}, \bibinfo{person}{Hao Liu}, \bibinfo{person}{Tao Tao},
  \bibinfo{person}{Naiqiang Tan}, {and} \bibinfo{person}{Hui Xiong}.}
  \bibinfo{year}{2024}\natexlab{}.
\newblock \showarticletitle{BigST: Linear Complexity Spatio-Temporal Graph
  Neural Network for Traffic Forecasting on Large-Scale Road Networks}.
\newblock \bibinfo{journal}{\emph{Proceedings of the VLDB Endowment}}
  (\bibinfo{year}{2024}), \bibinfo{pages}{1081--1090}.
\newblock


\bibitem[Horn et~al\mbox{.}(2020)]%
        {horn2020set}
\bibfield{author}{\bibinfo{person}{Max Horn}, \bibinfo{person}{Michael Moor},
  \bibinfo{person}{Christian Bock}, \bibinfo{person}{Bastian Rieck}, {and}
  \bibinfo{person}{Karsten Borgwardt}.} \bibinfo{year}{2020}\natexlab{}.
\newblock \showarticletitle{Set functions for time series}. In
  \bibinfo{booktitle}{\emph{International Conference on Machine Learning}}.
  \bibinfo{pages}{4353--4363}.
\newblock


\bibitem[Ji et~al\mbox{.}(2023)]%
        {ji2023spatio}
\bibfield{author}{\bibinfo{person}{Jiahao Ji}, \bibinfo{person}{Jingyuan Wang},
  \bibinfo{person}{Chao Huang}, \bibinfo{person}{Junjie Wu},
  \bibinfo{person}{Boren Xu}, \bibinfo{person}{Zhenhe Wu},
  \bibinfo{person}{Junbo Zhang}, {and} \bibinfo{person}{Yu Zheng}.}
  \bibinfo{year}{2023}\natexlab{}.
\newblock \showarticletitle{Spatio-temporal self-supervised learning for
  traffic flow prediction}. In \bibinfo{booktitle}{\emph{Proceedings of the
  AAAI conference on artificial intelligence}}, Vol.~\bibinfo{volume}{37}.
  \bibinfo{pages}{4356--4364}.
\newblock


\bibitem[Jiang et~al\mbox{.}(2023)]%
        {pdformer2023}
\bibfield{author}{\bibinfo{person}{Jiawei Jiang}, \bibinfo{person}{Chengkai
  Han}, \bibinfo{person}{Wayne~Xin Zhao}, {and} \bibinfo{person}{Jingyuan
  Wang}.} \bibinfo{year}{2023}\natexlab{}.
\newblock \showarticletitle{PDFormer: Propagation Delay-aware Dynamic
  Long-range Transformer for Traffic Flow Prediction}. In
  \bibinfo{booktitle}{\emph{{Proceedings of the AAAI conference on artificial
  intelligence}}}. \bibinfo{pages}{4365--4373}.
\newblock


\bibitem[Jiang and Luo(2022)]%
        {jiang2022graph}
\bibfield{author}{\bibinfo{person}{Weiwei Jiang} {and} \bibinfo{person}{Jiayun
  Luo}.} \bibinfo{year}{2022}\natexlab{}.
\newblock \showarticletitle{Graph neural network for traffic forecasting: A
  survey}.
\newblock \bibinfo{journal}{\emph{Expert Systems with Applications}}
  (\bibinfo{year}{2022}), \bibinfo{pages}{117921}.
\newblock


\bibitem[Kipf and Welling(2017)]%
        {kipf2017semi}
\bibfield{author}{\bibinfo{person}{Thomas~N. Kipf} {and} \bibinfo{person}{Max
  Welling}.} \bibinfo{year}{2017}\natexlab{}.
\newblock \showarticletitle{Semi-Supervised Classification with Graph
  Convolutional Networks}. In \bibinfo{booktitle}{\emph{International
  Conference on Learning Representations, {ICLR}}}.
\newblock


\bibitem[Kohli et~al\mbox{.}(2016)]%
        {kohli2016traffic}
\bibfield{author}{\bibinfo{person}{Serbjeet Kohli},
  \bibinfo{person}{Steer~Davies Gleave}, {and} \bibinfo{person}{Luis
  Willumsen}.} \bibinfo{year}{2016}\natexlab{}.
\newblock \showarticletitle{Traffic Forecasting and Autonomous Vehicles}. In
  \bibinfo{booktitle}{\emph{2016 European Transprot Conference, Barcelona}}.
  \bibinfo{pages}{5--7}.
\newblock


\bibitem[Lan et~al\mbox{.}(2022)]%
        {lan2022dstagnn}
\bibfield{author}{\bibinfo{person}{Shiyong Lan}, \bibinfo{person}{Yitong Ma},
  \bibinfo{person}{Weikang Huang}, \bibinfo{person}{Wenwu Wang},
  \bibinfo{person}{Hongyu Yang}, {and} \bibinfo{person}{Pyang Li}.}
  \bibinfo{year}{2022}\natexlab{}.
\newblock \showarticletitle{Dstagnn: Dynamic spatial-temporal aware graph
  neural network for traffic flow forecasting}. In
  \bibinfo{booktitle}{\emph{International Conference on Machine Learning}}.
  \bibinfo{pages}{11906--11917}.
\newblock


\bibitem[Lana et~al\mbox{.}(2018)]%
        {lana2018road}
\bibfield{author}{\bibinfo{person}{Ibai Lana}, \bibinfo{person}{Javier
  Del~Ser}, \bibinfo{person}{Manuel Velez}, {and} \bibinfo{person}{Eleni~I
  Vlahogianni}.} \bibinfo{year}{2018}\natexlab{}.
\newblock \showarticletitle{Road traffic forecasting: Recent advances and new
  challenges}.
\newblock \bibinfo{journal}{\emph{IEEE Intelligent Transportation Systems
  Magazine}} \bibinfo{volume}{10}, \bibinfo{number}{2} (\bibinfo{year}{2018}),
  \bibinfo{pages}{93--109}.
\newblock


\bibitem[LeCun et~al\mbox{.}(2015)]%
        {lecun2015deep}
\bibfield{author}{\bibinfo{person}{Yann LeCun}, \bibinfo{person}{Yoshua
  Bengio}, {and} \bibinfo{person}{Geoffrey Hinton}.}
  \bibinfo{year}{2015}\natexlab{}.
\newblock \showarticletitle{Deep learning}.
\newblock \bibinfo{journal}{\emph{nature}} (\bibinfo{year}{2015}),
  \bibinfo{pages}{436--444}.
\newblock


\bibitem[Li et~al\mbox{.}(2018)]%
        {li2018diffusion}
\bibfield{author}{\bibinfo{person}{Yaguang Li}, \bibinfo{person}{Rose Yu},
  \bibinfo{person}{Cyrus Shahabi}, {and} \bibinfo{person}{Yan Liu}.}
  \bibinfo{year}{2018}\natexlab{}.
\newblock \showarticletitle{Diffusion Convolutional Recurrent Neural Network:
  Data-Driven Traffic Forecasting}. In \bibinfo{booktitle}{\emph{International
  Conference on Learning Representations}}.
\newblock


\bibitem[Lipton et~al\mbox{.}(2016)]%
        {lipton2016directly}
\bibfield{author}{\bibinfo{person}{Zachary~C Lipton}, \bibinfo{person}{David
  Kale}, {and} \bibinfo{person}{Randall Wetzel}.}
  \bibinfo{year}{2016}\natexlab{}.
\newblock \showarticletitle{Directly modeling missing data in sequences with
  rnns: Improved classification of clinical time series}. In
  \bibinfo{booktitle}{\emph{Machine learning for healthcare conference}}.
  \bibinfo{pages}{253--270}.
\newblock


\bibitem[Liu et~al\mbox{.}(2023b)]%
        {liu2023robust}
\bibfield{author}{\bibinfo{person}{Fan Liu}, \bibinfo{person}{Weijia Zhang},
  {and} \bibinfo{person}{Hao Liu}.} \bibinfo{year}{2023}\natexlab{b}.
\newblock \showarticletitle{Robust Spatiotemporal Traffic Forecasting with
  Reinforced Dynamic Adversarial Training}. In
  \bibinfo{booktitle}{\emph{Proceedings of the 29th ACM SIGKDD Conference on
  Knowledge Discovery and Data Mining}}. \bibinfo{pages}{1417–1428}.
\newblock


\bibitem[Liu et~al\mbox{.}(2023a)]%
        {liu2023spatio}
\bibfield{author}{\bibinfo{person}{Hangchen Liu}, \bibinfo{person}{Zheng Dong},
  \bibinfo{person}{Renhe Jiang}, \bibinfo{person}{Jiewen Deng},
  \bibinfo{person}{Jinliang Deng}, \bibinfo{person}{Quanjun Chen}, {and}
  \bibinfo{person}{Xuan Song}.} \bibinfo{year}{2023}\natexlab{a}.
\newblock \showarticletitle{Spatio-temporal adaptive embedding makes vanilla
  transformer sota for traffic forecasting}. In
  \bibinfo{booktitle}{\emph{Proceedings of the 32nd ACM international
  conference on information and knowledge management}}.
  \bibinfo{pages}{4125--4129}.
\newblock


\bibitem[Marisca et~al\mbox{.}(2022)]%
        {mariscalearning}
\bibfield{author}{\bibinfo{person}{Ivan Marisca}, \bibinfo{person}{Andrea
  Cini}, {and} \bibinfo{person}{Cesare Alippi}.}
  \bibinfo{year}{2022}\natexlab{}.
\newblock \showarticletitle{Learning to Reconstruct Missing Data from
  Spatiotemporal Graphs with Sparse Observations}. In
  \bibinfo{booktitle}{\emph{Advances in Neural Information Processing
  Systems}}.
\newblock


\bibitem[Moreno-Pino et~al\mbox{.}(2023)]%
        {moreno2023deep}
\bibfield{author}{\bibinfo{person}{Fernando Moreno-Pino},
  \bibinfo{person}{Pablo~M Olmos}, {and} \bibinfo{person}{Antonio
  Art{\'e}s-Rodr{\'\i}guez}.} \bibinfo{year}{2023}\natexlab{}.
\newblock \showarticletitle{Deep autoregressive models with spectral
  attention}.
\newblock \bibinfo{journal}{\emph{Pattern Recognition}} (\bibinfo{year}{2023}),
  \bibinfo{pages}{109014}.
\newblock


\bibitem[Neil et~al\mbox{.}(2016)]%
        {neil2016phased}
\bibfield{author}{\bibinfo{person}{Daniel Neil}, \bibinfo{person}{Michael
  Pfeiffer}, {and} \bibinfo{person}{Shih-Chii Liu}.}
  \bibinfo{year}{2016}\natexlab{}.
\newblock \showarticletitle{Phased LSTM: accelerating recurrent network
  training for long or event-based sequences}. In
  \bibinfo{booktitle}{\emph{Proceedings of the 30th International Conference on
  Neural Information Processing Systems}}. \bibinfo{pages}{3889--3897}.
\newblock


\bibitem[Rahman and Hasan(2023)]%
        {rahman2023deep}
\bibfield{author}{\bibinfo{person}{Rezaur Rahman} {and} \bibinfo{person}{Samiul
  Hasan}.} \bibinfo{year}{2023}\natexlab{}.
\newblock \showarticletitle{A deep learning approach for network-wide dynamic
  traffic prediction during hurricane evacuation}.
\newblock \bibinfo{journal}{\emph{Transportation Research Part C: Emerging
  Technologies}}  \bibinfo{volume}{152} (\bibinfo{year}{2023}),
  \bibinfo{pages}{104126}.
\newblock


\bibitem[Rubanova et~al\mbox{.}(2019)]%
        {rubanova2019latent}
\bibfield{author}{\bibinfo{person}{Yulia Rubanova}, \bibinfo{person}{Ricky~TQ
  Chen}, {and} \bibinfo{person}{David Duvenaud}.}
  \bibinfo{year}{2019}\natexlab{}.
\newblock \showarticletitle{Latent ODEs for irregularly-sampled time series}.
  In \bibinfo{booktitle}{\emph{Proceedings of the 33rd International Conference
  on Neural Information Processing Systems}}. \bibinfo{pages}{5320--5330}.
\newblock


\bibitem[Schirmer et~al\mbox{.}(2022)]%
        {schirmer2022modeling}
\bibfield{author}{\bibinfo{person}{Mona Schirmer}, \bibinfo{person}{Mazin
  Eltayeb}, \bibinfo{person}{Stefan Lessmann}, {and} \bibinfo{person}{Maja
  Rudolph}.} \bibinfo{year}{2022}\natexlab{}.
\newblock \showarticletitle{Modeling irregular time series with continuous
  recurrent units}. In \bibinfo{booktitle}{\emph{International Conference on
  Machine Learning}}. \bibinfo{pages}{19388--19405}.
\newblock


\bibitem[Shao et~al\mbox{.}(2022a)]%
        {shao2022pre}
\bibfield{author}{\bibinfo{person}{Zezhi Shao}, \bibinfo{person}{Zhao Zhang},
  \bibinfo{person}{Fei Wang}, {and} \bibinfo{person}{Yongjun Xu}.}
  \bibinfo{year}{2022}\natexlab{a}.
\newblock \showarticletitle{Pre-training enhanced spatial-temporal graph neural
  network for multivariate time series forecasting}. In
  \bibinfo{booktitle}{\emph{Proceedings of the 28th ACM SIGKDD Conference on
  Knowledge Discovery and Data Mining}}. \bibinfo{pages}{1567--1577}.
\newblock


\bibitem[Shao et~al\mbox{.}(2022b)]%
        {shao2022decoupled}
\bibfield{author}{\bibinfo{person}{Zezhi Shao}, \bibinfo{person}{Zhao Zhang},
  \bibinfo{person}{Wei Wei}, \bibinfo{person}{Fei Wang},
  \bibinfo{person}{Yongjun Xu}, \bibinfo{person}{Xin Cao}, {and}
  \bibinfo{person}{Christian~S Jensen}.} \bibinfo{year}{2022}\natexlab{b}.
\newblock \showarticletitle{Decoupled dynamic spatial-temporal graph neural
  network for traffic forecasting}.
\newblock \bibinfo{journal}{\emph{Proceedings of the VLDB Endowment}}
  (\bibinfo{year}{2022}), \bibinfo{pages}{2733--2746}.
\newblock


\bibitem[Shukla and Marlin(2021)]%
        {shukla2020multi}
\bibfield{author}{\bibinfo{person}{Satya~Narayan Shukla} {and}
  \bibinfo{person}{Benjamin Marlin}.} \bibinfo{year}{2021}\natexlab{}.
\newblock \showarticletitle{Multi-Time Attention Networks for Irregularly
  Sampled Time Series}. In \bibinfo{booktitle}{\emph{International Conference
  on Learning Representations}}.
\newblock


\bibitem[Shukla and Marlin(2020)]%
        {shukla2020survey}
\bibfield{author}{\bibinfo{person}{Satya~Narayan Shukla} {and}
  \bibinfo{person}{Benjamin~M Marlin}.} \bibinfo{year}{2020}\natexlab{}.
\newblock \showarticletitle{A survey on principles, models and methods for
  learning from irregularly sampled time series}.
\newblock \bibinfo{journal}{\emph{arXiv preprint arXiv:2012.00168}}
  (\bibinfo{year}{2020}).
\newblock


\bibitem[Sun et~al\mbox{.}(2023)]%
        {sun2023hierarchical}
\bibfield{author}{\bibinfo{person}{Qian Sun}, \bibinfo{person}{Le Zhang},
  \bibinfo{person}{Huan Yu}, \bibinfo{person}{Weijia Zhang},
  \bibinfo{person}{Yu Mei}, {and} \bibinfo{person}{Hui Xiong}.}
  \bibinfo{year}{2023}\natexlab{}.
\newblock \showarticletitle{Hierarchical reinforcement learning for dynamic
  autonomous vehicle navigation at intelligent intersections}. In
  \bibinfo{booktitle}{\emph{Proceedings of the 29th ACM SIGKDD Conference on
  Knowledge Discovery and Data Mining}}. \bibinfo{pages}{4852--4861}.
\newblock


\bibitem[Vaswani et~al\mbox{.}(2017)]%
        {vaswani2017attention}
\bibfield{author}{\bibinfo{person}{Ashish Vaswani}, \bibinfo{person}{Noam
  Shazeer}, \bibinfo{person}{Niki Parmar}, \bibinfo{person}{Jakob Uszkoreit},
  \bibinfo{person}{Llion Jones}, \bibinfo{person}{Aidan~N Gomez},
  \bibinfo{person}{{\L}ukasz Kaiser}, {and} \bibinfo{person}{Illia
  Polosukhin}.} \bibinfo{year}{2017}\natexlab{}.
\newblock \showarticletitle{Attention is all you need}.
\newblock \bibinfo{journal}{\emph{Advances in neural information processing
  systems}}  \bibinfo{volume}{30} (\bibinfo{year}{2017}).
\newblock


\bibitem[Veli{\v{c}}kovi{\'c} et~al\mbox{.}(2018)]%
        {velivckovic2018graph}
\bibfield{author}{\bibinfo{person}{Petar Veli{\v{c}}kovi{\'c}},
  \bibinfo{person}{Guillem Cucurull}, \bibinfo{person}{Arantxa Casanova},
  \bibinfo{person}{Adriana Romero}, \bibinfo{person}{Pietro Li{\`o}}, {and}
  \bibinfo{person}{Yoshua Bengio}.} \bibinfo{year}{2018}\natexlab{}.
\newblock \showarticletitle{Graph Attention Networks}. In
  \bibinfo{booktitle}{\emph{International Conference on Learning
  Representations}}.
\newblock


\bibitem[Wang et~al\mbox{.}(2016)]%
        {wang2016dynamic}
\bibfield{author}{\bibinfo{person}{Chen Wang}, \bibinfo{person}{Bertrand
  David}, \bibinfo{person}{Ren{\'e} Chalon}, {and} \bibinfo{person}{Chuantao
  Yin}.} \bibinfo{year}{2016}\natexlab{}.
\newblock \showarticletitle{Dynamic road lane management study: A Smart City
  application}.
\newblock \bibinfo{journal}{\emph{Transportation research part E: logistics and
  transportation review}}  \bibinfo{volume}{89} (\bibinfo{year}{2016}),
  \bibinfo{pages}{272--287}.
\newblock


\bibitem[Wang et~al\mbox{.}(2018)]%
        {wang2018review}
\bibfield{author}{\bibinfo{person}{Yizhe Wang}, \bibinfo{person}{Xiaoguang
  Yang}, \bibinfo{person}{Hailun Liang}, \bibinfo{person}{Yangdong Liu},
  {et~al\mbox{.}}} \bibinfo{year}{2018}\natexlab{}.
\newblock \showarticletitle{A review of the self-adaptive traffic signal
  control system based on future traffic environment}.
\newblock \bibinfo{journal}{\emph{Journal of Advanced Transportation}}
  \bibinfo{volume}{2018} (\bibinfo{year}{2018}).
\newblock


\bibitem[Wei et~al\mbox{.}(2019)]%
        {wei2019survey}
\bibfield{author}{\bibinfo{person}{Hua Wei}, \bibinfo{person}{Guanjie Zheng},
  \bibinfo{person}{Vikash Gayah}, {and} \bibinfo{person}{Zhenhui Li}.}
  \bibinfo{year}{2019}\natexlab{}.
\newblock \showarticletitle{A survey on traffic signal control methods}.
\newblock \bibinfo{journal}{\emph{arXiv preprint arXiv:1904.08117}}
  (\bibinfo{year}{2019}).
\newblock


\bibitem[Wei et~al\mbox{.}(2023)]%
        {wei2023compatible}
\bibfield{author}{\bibinfo{person}{Yuxi Wei}, \bibinfo{person}{Juntong Peng},
  \bibinfo{person}{Tong He}, \bibinfo{person}{Chenxin Xu},
  \bibinfo{person}{Jian Zhang}, \bibinfo{person}{Shirui Pan}, {and}
  \bibinfo{person}{Siheng Chen}.} \bibinfo{year}{2023}\natexlab{}.
\newblock \showarticletitle{Compatible transformer for irregularly sampled
  multivariate time series}. In \bibinfo{booktitle}{\emph{2023 IEEE
  International Conference on Data Mining (ICDM)}}. IEEE,
  \bibinfo{pages}{1409--1414}.
\newblock


\bibitem[Wu et~al\mbox{.}(2020)]%
        {wu2020comprehensive}
\bibfield{author}{\bibinfo{person}{Zonghan Wu}, \bibinfo{person}{Shirui Pan},
  \bibinfo{person}{Fengwen Chen}, \bibinfo{person}{Guodong Long},
  \bibinfo{person}{Chengqi Zhang}, {and} \bibinfo{person}{S~Yu Philip}.}
  \bibinfo{year}{2020}\natexlab{}.
\newblock \showarticletitle{A comprehensive survey on graph neural networks}.
\newblock \bibinfo{journal}{\emph{IEEE transactions on neural networks and
  learning systems}} (\bibinfo{year}{2020}), \bibinfo{pages}{4--24}.
\newblock


\bibitem[Wu et~al\mbox{.}(2019)]%
        {wu2019graph}
\bibfield{author}{\bibinfo{person}{Zonghan Wu}, \bibinfo{person}{Shirui Pan},
  \bibinfo{person}{Guodong Long}, \bibinfo{person}{Jing Jiang}, {and}
  \bibinfo{person}{Chengqi Zhang}.} \bibinfo{year}{2019}\natexlab{}.
\newblock \showarticletitle{Graph wavenet for deep spatial-temporal graph
  modeling}. In \bibinfo{booktitle}{\emph{Proceedings of the International
  Joint Conference on Artificial Intelligence}}. \bibinfo{pages}{1907--1913}.
\newblock


\bibitem[Yao et~al\mbox{.}(2018)]%
        {yao2018deep}
\bibfield{author}{\bibinfo{person}{Huaxiu Yao}, \bibinfo{person}{Fei Wu},
  \bibinfo{person}{Jintao Ke}, \bibinfo{person}{Xianfeng Tang},
  \bibinfo{person}{Yitian Jia}, \bibinfo{person}{Siyu Lu},
  \bibinfo{person}{Pinghua Gong}, \bibinfo{person}{Zhenhui Li},
  \bibinfo{person}{Jieping Ye}, {and} \bibinfo{person}{Didi Chuxing}.}
  \bibinfo{year}{2018}\natexlab{}.
\newblock \showarticletitle{Deep multi-view spatial-temporal network for taxi
  demand prediction}. In \bibinfo{booktitle}{\emph{Proceedings of the AAAI
  conference on artificial intelligence}}. \bibinfo{pages}{2588--2595}.
\newblock


\bibitem[Yi et~al\mbox{.}(2023)]%
        {yi2023fouriergnn}
\bibfield{author}{\bibinfo{person}{Kun Yi}, \bibinfo{person}{Qi Zhang},
  \bibinfo{person}{Wei Fan}, \bibinfo{person}{Hui He}, \bibinfo{person}{Liang
  Hu}, \bibinfo{person}{Pengyang Wang}, \bibinfo{person}{Ning An},
  \bibinfo{person}{Longbing Cao}, {and} \bibinfo{person}{Zhendong Niu}.}
  \bibinfo{year}{2023}\natexlab{}.
\newblock \showarticletitle{FourierGNN: Rethinking multivariate time series
  forecasting from a pure graph perspective}.
\newblock \bibinfo{journal}{\emph{Advances in Neural Information Processing
  Systems}}  \bibinfo{volume}{36} (\bibinfo{year}{2023}).
\newblock


\bibitem[Yu et~al\mbox{.}(2018)]%
        {yu2018spatio}
\bibfield{author}{\bibinfo{person}{Bing Yu}, \bibinfo{person}{Haoteng Yin},
  {and} \bibinfo{person}{Zhanxing Zhu}.} \bibinfo{year}{2018}\natexlab{}.
\newblock \showarticletitle{Spatio-temporal graph convolutional networks: a
  deep learning framework for traffic forecasting}. In
  \bibinfo{booktitle}{\emph{Proceedings of the 27th International Joint
  Conference on Artificial Intelligence}}. \bibinfo{pages}{3634--3640}.
\newblock


\bibitem[Yuan et~al\mbox{.}(2020)]%
        {yuan2020spatio}
\bibfield{author}{\bibinfo{person}{Zixuan Yuan}, \bibinfo{person}{Hao Liu},
  \bibinfo{person}{Yanchi Liu}, \bibinfo{person}{Denghui Zhang},
  \bibinfo{person}{Fei Yi}, \bibinfo{person}{Nengjun Zhu}, {and}
  \bibinfo{person}{Hui Xiong}.} \bibinfo{year}{2020}\natexlab{}.
\newblock \showarticletitle{Spatio-temporal dual graph attention network for
  query-poi matching}. In \bibinfo{booktitle}{\emph{Proceedings of the 43rd
  international ACM SIGIR conference on research and development in information
  retrieval}}. \bibinfo{pages}{629--638}.
\newblock


\bibitem[Zhang et~al\mbox{.}(2023)]%
        {warpformer2023}
\bibfield{author}{\bibinfo{person}{Jiawen Zhang}, \bibinfo{person}{Shun Zheng},
  \bibinfo{person}{Wei Cao}, \bibinfo{person}{Jiang Bian}, {and}
  \bibinfo{person}{Jia Li}.} \bibinfo{year}{2023}\natexlab{}.
\newblock \showarticletitle{Warpformer: A Multi-Scale Modeling Approach for
  Irregular Clinical Time Series}. In \bibinfo{booktitle}{\emph{Proceedings of
  the 29th ACM SIGKDD Conference on Knowledge Discovery and Data Mining}}.
  \bibinfo{pages}{3273–3285}.
\newblock


\bibitem[Zhang et~al\mbox{.}(2017)]%
        {zhang2017deep}
\bibfield{author}{\bibinfo{person}{Junbo Zhang}, \bibinfo{person}{Yu Zheng},
  {and} \bibinfo{person}{Dekang Qi}.} \bibinfo{year}{2017}\natexlab{}.
\newblock \showarticletitle{Deep spatio-temporal residual networks for citywide
  crowd flows prediction}. In \bibinfo{booktitle}{\emph{Proceedings of the AAAI
  conference on artificial intelligence}}. \bibinfo{pages}{1655--1661}.
\newblock


\bibitem[Zhang et~al\mbox{.}(2022a)]%
        {zhang2022multi}
\bibfield{author}{\bibinfo{person}{Weijia Zhang}, \bibinfo{person}{Hao Liu},
  \bibinfo{person}{Jindong Han}, \bibinfo{person}{Yong Ge}, {and}
  \bibinfo{person}{Hui Xiong}.} \bibinfo{year}{2022}\natexlab{a}.
\newblock \showarticletitle{Multi-agent graph convolutional reinforcement
  learning for dynamic electric vehicle charging pricing}. In
  \bibinfo{booktitle}{\emph{Proceedings of the 28th ACM SIGKDD Conference on
  Knowledge Discovery and Data Mining}}. \bibinfo{pages}{2471--2481}.
\newblock


\bibitem[Zhang et~al\mbox{.}(2020)]%
        {zhang2020semi}
\bibfield{author}{\bibinfo{person}{Weijia Zhang}, \bibinfo{person}{Hao Liu},
  \bibinfo{person}{Yanchi Liu}, \bibinfo{person}{Jingbo Zhou}, {and}
  \bibinfo{person}{Hui Xiong}.} \bibinfo{year}{2020}\natexlab{}.
\newblock \showarticletitle{Semi-supervised hierarchical recurrent graph neural
  network for city-wide parking availability prediction}. In
  \bibinfo{booktitle}{\emph{Proceedings of the AAAI Conference on Artificial
  Intelligence}}. \bibinfo{pages}{1186--1193}.
\newblock


\bibitem[Zhang et~al\mbox{.}(2022b)]%
        {zhang2021graph}
\bibfield{author}{\bibinfo{person}{Xiang Zhang}, \bibinfo{person}{Marko Zeman},
  \bibinfo{person}{Theodoros Tsiligkaridis}, {and} \bibinfo{person}{Marinka
  Zitnik}.} \bibinfo{year}{2022}\natexlab{b}.
\newblock \showarticletitle{Graph-Guided Network for Irregularly Sampled
  Multivariate Time Series}. In \bibinfo{booktitle}{\emph{International
  Conference on Learning Representations}}.
\newblock


\bibitem[Zheng et~al\mbox{.}(2020)]%
        {zheng2020gman}
\bibfield{author}{\bibinfo{person}{Chuanpan Zheng}, \bibinfo{person}{Xiaoliang
  Fan}, \bibinfo{person}{Cheng Wang}, {and} \bibinfo{person}{Jianzhong Qi}.}
  \bibinfo{year}{2020}\natexlab{}.
\newblock \showarticletitle{Gman: A graph multi-attention network for traffic
  prediction}. In \bibinfo{booktitle}{\emph{Proceedings of the AAAI conference
  on artificial intelligence}}. \bibinfo{pages}{1234--1241}.
\newblock


\end{thebibliography}

\appendix

\section{Supplementary Experiments}

\begin{figure}[b]
  \centering
  \hspace{-2mm}
  \subfigure[{\zhuzhou.}]{
    \includegraphics[width=0.505\columnwidth]{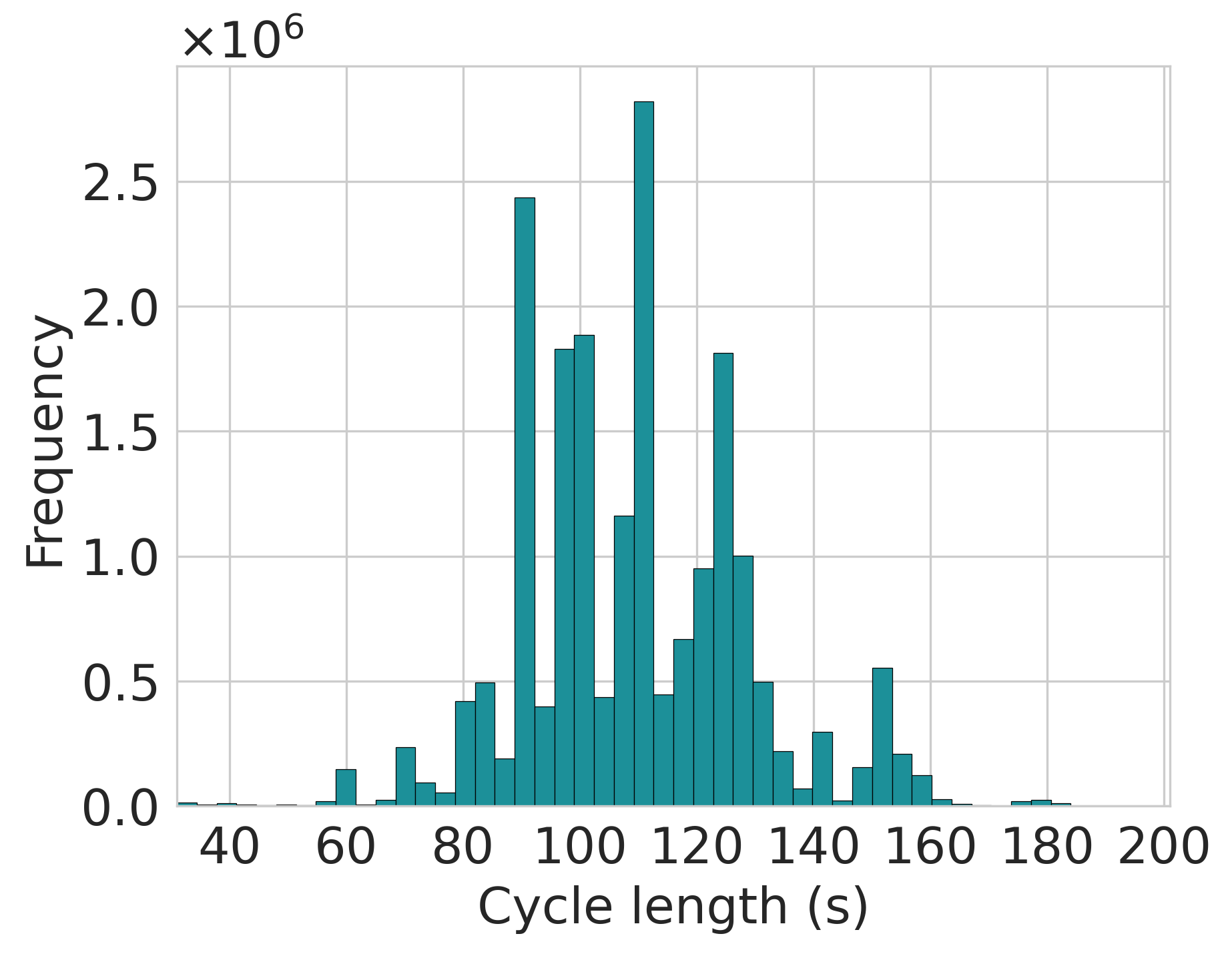}}
  \subfigure[{\baoding.}]{
    \includegraphics[width=0.505\columnwidth]{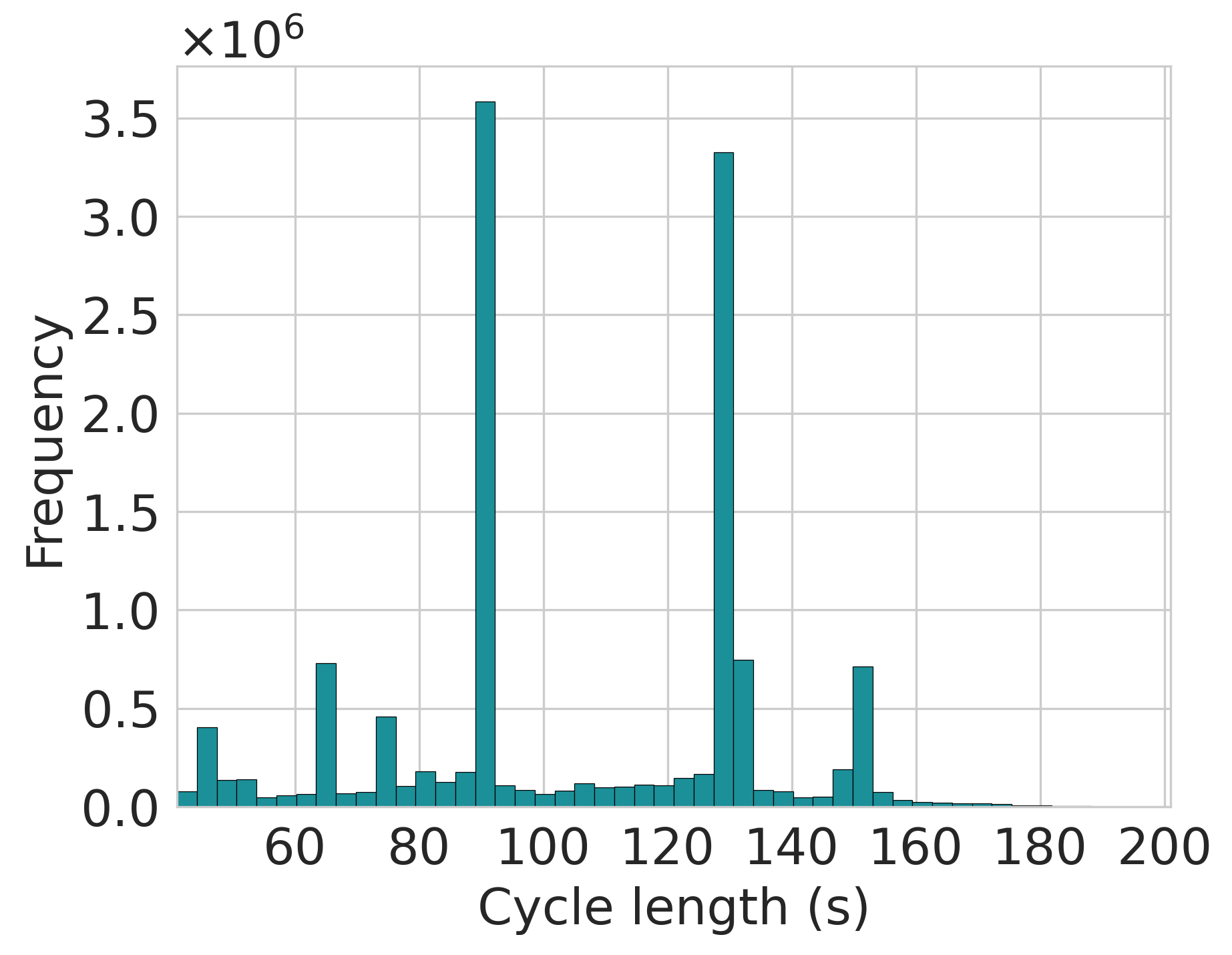}}
  \caption{Overall distributions of traffic signal cycle lengths.} 
  \label{fig:cycle_dist}
\end{figure}

\subsection{\textbf{Data Description and Analysis}} \label{app:dataset}

\subsubsection{\textbf{Datasets Description}} 
The statistics of the datasets are summarized
in \tabref{table:dataset}. Specifically, there are total 19,824,504 and 13,093,975 traffic state measurements on \zhuzhou and \baoding, and the missing period ratios of the two datasets are $44.2\%$ and $27.2\%$, respectively. 
Each measurement includes information about the beginning and end timestamps and cycle length of the traffic signal cycle, as well as the lane's traffic flow, \rev{\ie the number of vehicles passing through the camera of lane}, during the signal cycle.
Besides, \zhuzhou has $620$ lanes with sensors and ranges from July 20, 2022 to October 2, 2022. \baoding has $264$ lanes with sensors and ranges from December 1, 2021 to February 25, 2022.
The average, maximal ground truth sequence length to be predicted for the future one hour is 57, 213 on \zhuzhou, and 64, 155 on \baoding.

\subsubsection{\textbf{Datasets Analysis}} 
The overall distributions of traffic signal cycle lengths on two datasets are depicted in \figref{fig:cycle_dist}, where we can observe the cycle lengths can significantly vary from around 40 to 200 seconds on both datasets, indicating the pronounced irregularity within time series, and \baoding has a denser cycle length distribution than \zhuzhou.

\begin{figure*}[tb]
  \centering
  \subfigure[{Cycle lengths on \zhuzhou.}]{
    \includegraphics[width=0.505\columnwidth]{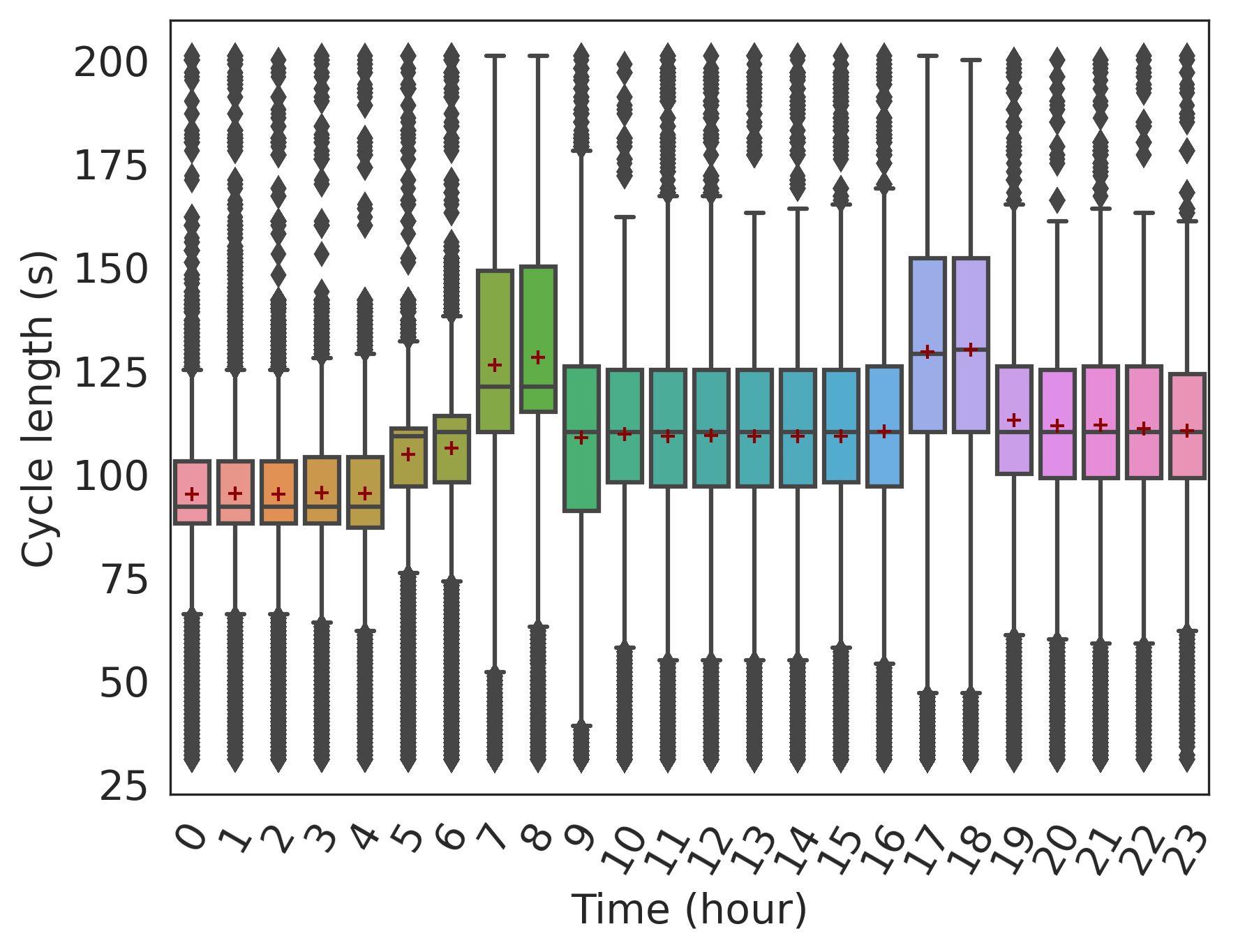}}
  \subfigure[{Cycle lengths on \baoding.}]{
    \includegraphics[width=0.505\columnwidth]{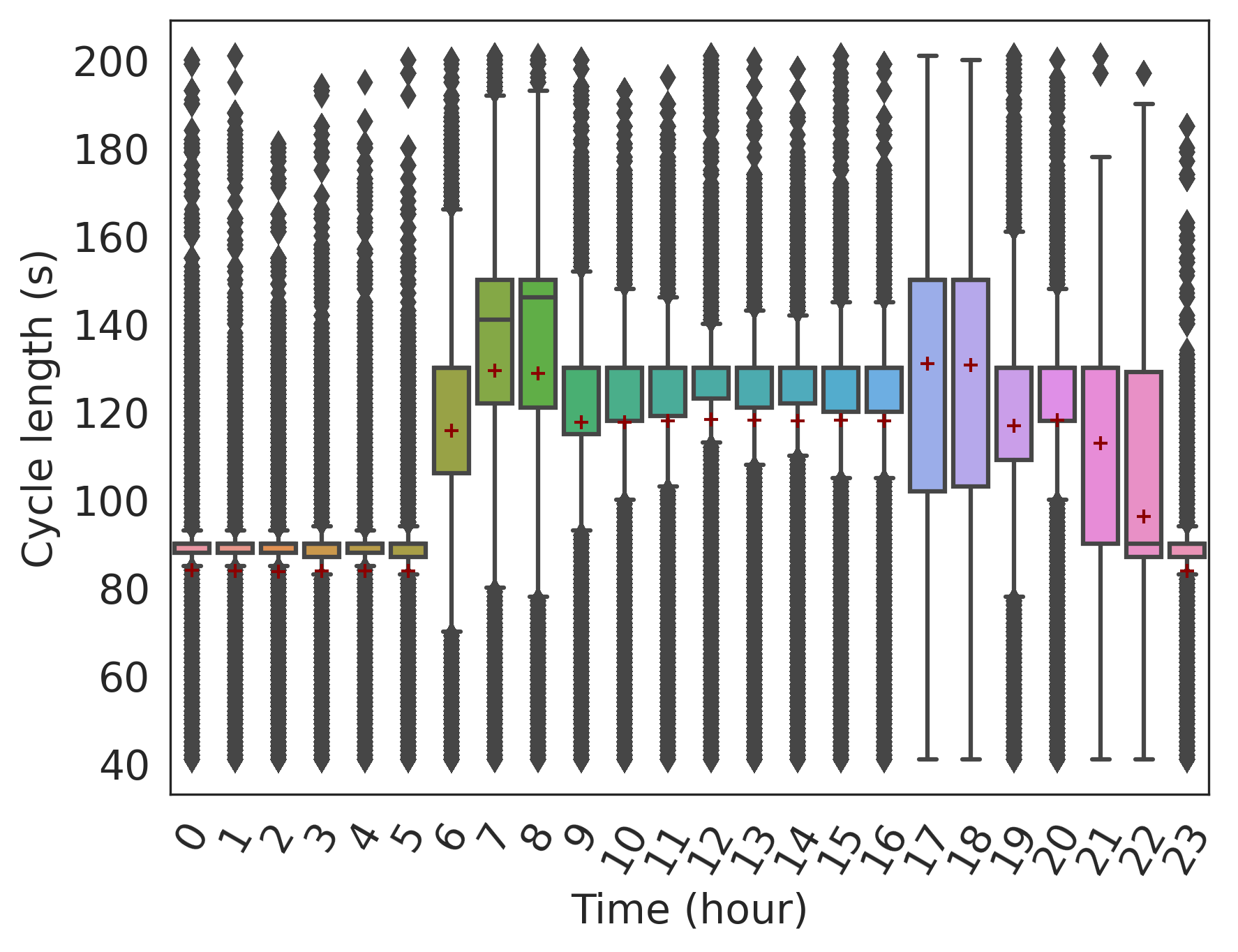}}
  \subfigure[{Traffic flows on \zhuzhou.}]{
    \includegraphics[width=0.5\columnwidth]{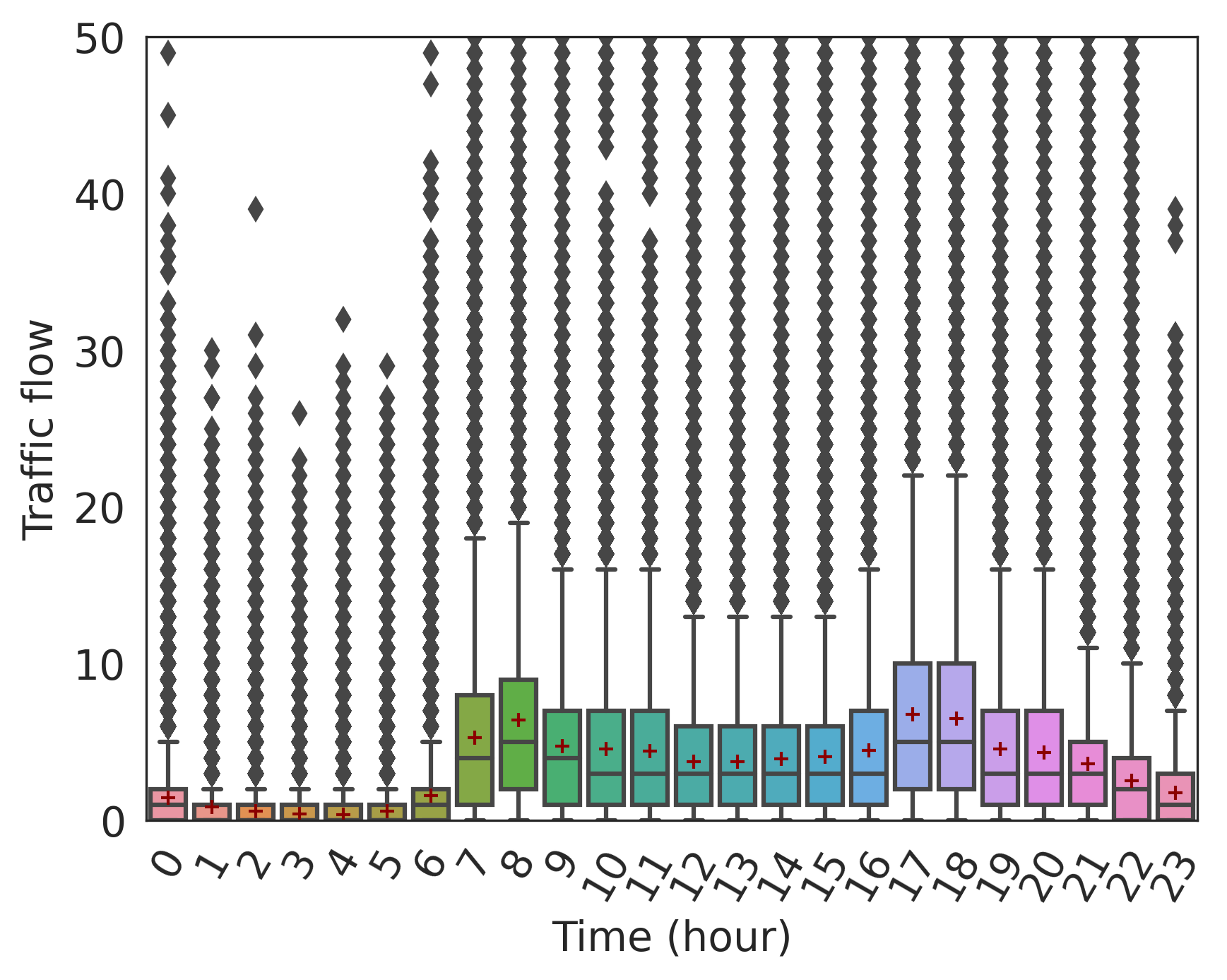}}
  \subfigure[{Traffic flows on \baoding.}]{
    \includegraphics[width=0.5\columnwidth]{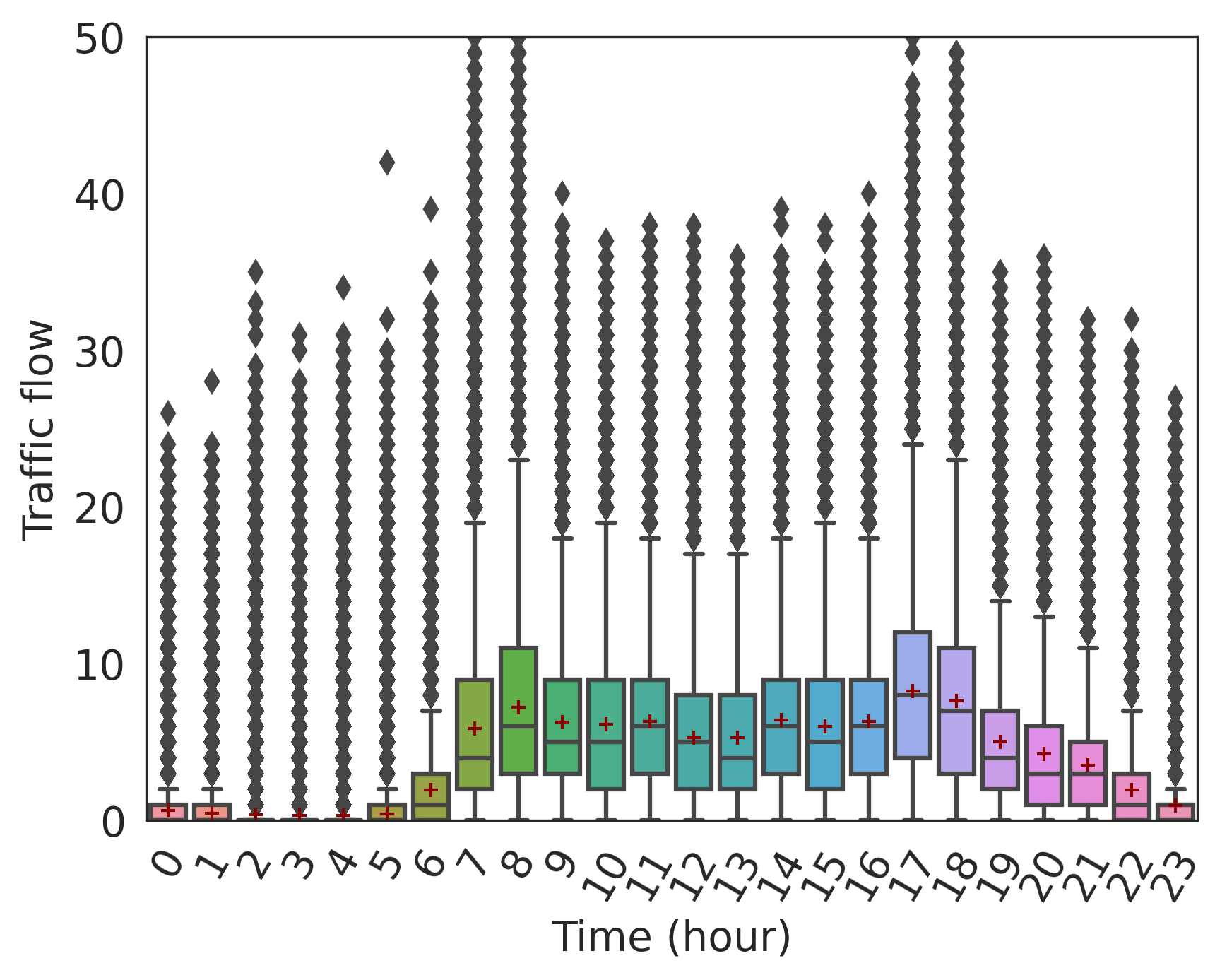}}
  \vspace{-3mm}
  \caption{Temporal distributions of traffic signal cycle lengths and traffic flows across time. `+' denotes mean of the box plot.} 
  \vspace{-2mm}
  \label{fig:box_over_time}
\end{figure*}

Besides, \figref{fig:box_over_time} illustrates temporal distributions of traffic signal cycle lengths and traffic flows across different hours on both datasets. 
We can observe cycle length and traffic flow consistently exhibit higher values during the daytime periods compared to overnight periods. Moreover, they display similar peak patterns during the morning and evening rush hours and tend to vary in a positively correlated manner. 

\begin{figure*}[tb]
  \centering
  \subfigure[{Distributions of traffic flows across different cycle lengths on \zhuzhou.}]{
    \includegraphics[width=0.5\columnwidth]{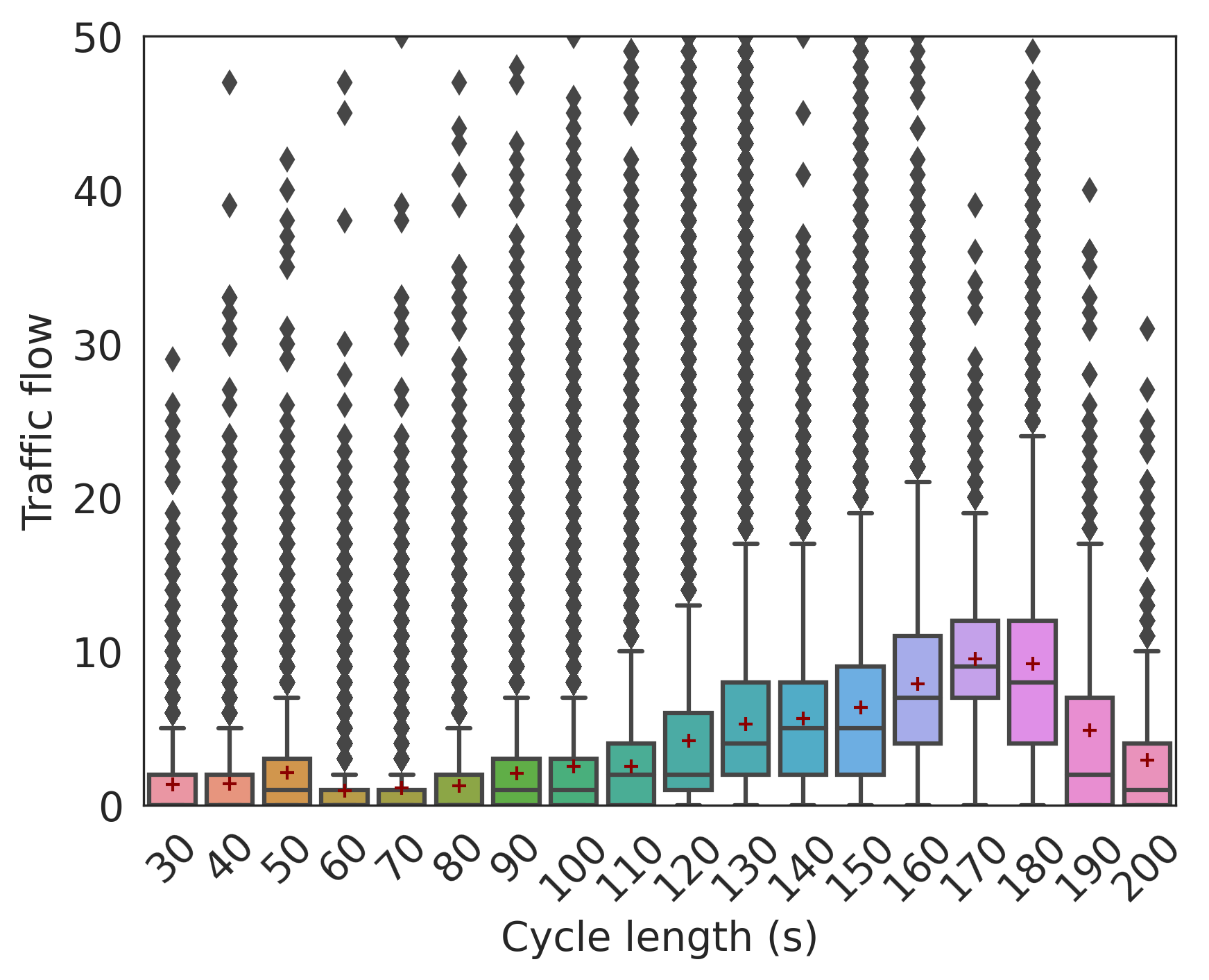}}
  \subfigure[{Distributions of traffic flows across different cycle lengths on \baoding.}]{
    \includegraphics[width=0.5\columnwidth]{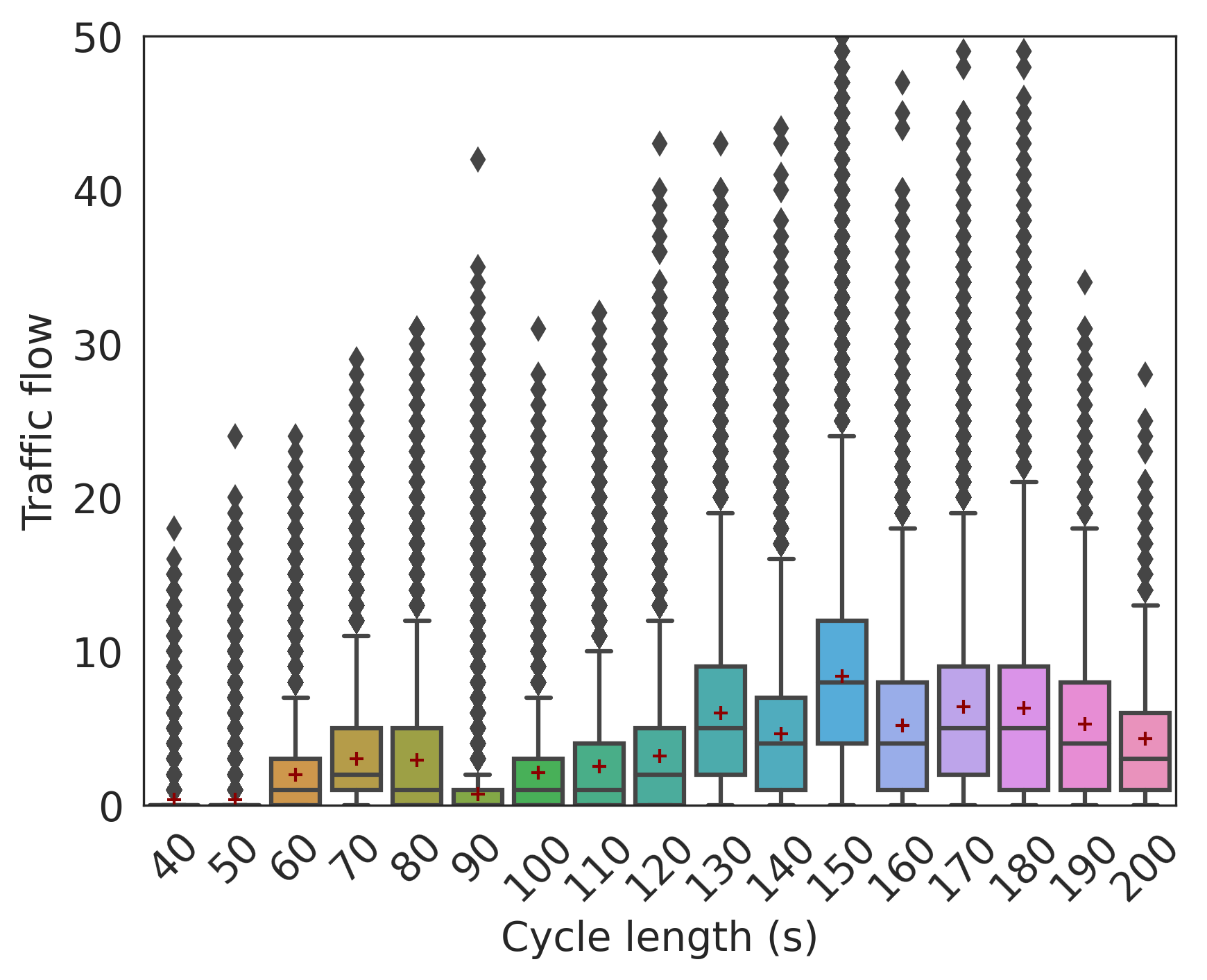}}
  \subfigure[{Distributions of cycle lengths across distinct traffic flows on \zhuzhou.}]{
    \includegraphics[width=0.50\columnwidth]{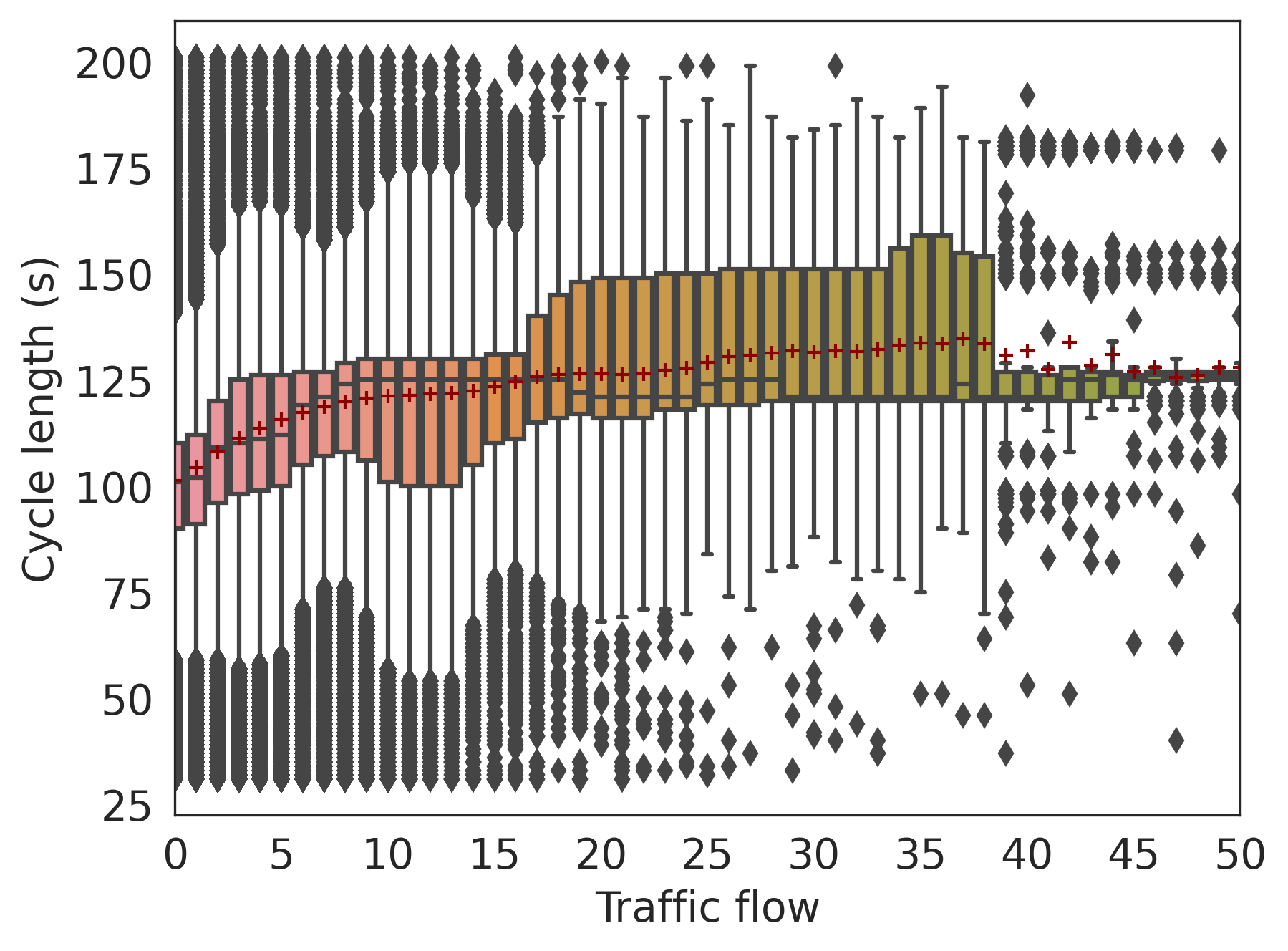}}
  \subfigure[{Distributions of cycle lengths across distinct traffic flows on \baoding.}]{
    \includegraphics[width=0.50\columnwidth]{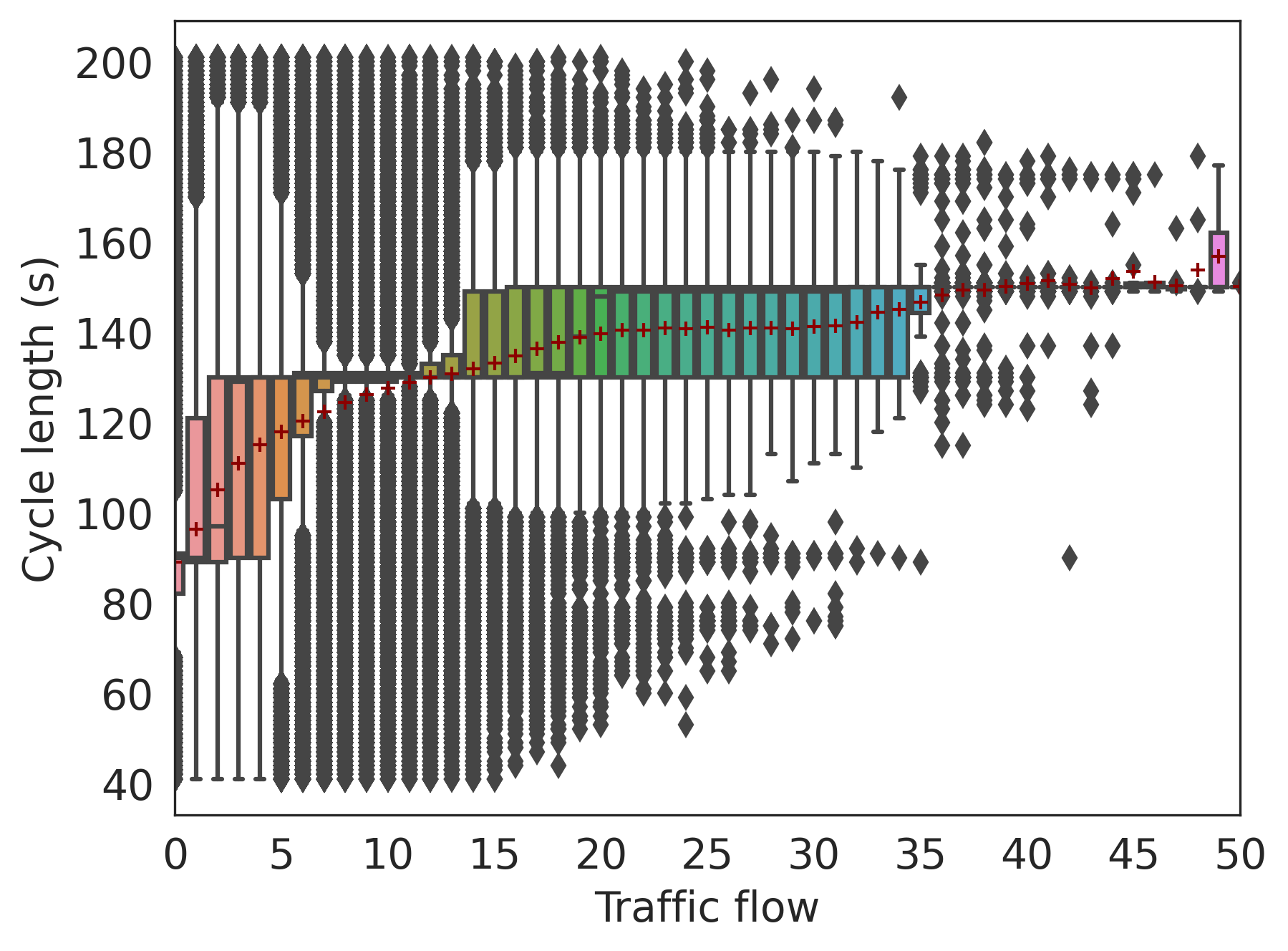}}
  \vspace{-3mm}
  \caption{Correlations between traffic flow and traffic signal cycle length. `+' denotes mean of the box plot.} 
  \vspace{-2mm}
  \label{fig:box_flow_cycle}
\end{figure*}

To further investigate the correlations between these two traffic states, we illustrate the variations in traffic flow distributions across different cycle lengths and vice versa in \figref{fig:box_flow_cycle}. As can be seen in \figref{fig:box_flow_cycle}(a) and \figref{fig:box_flow_cycle}(b), traffic flow maintains an upward trend at first along with the increase of cycle length. A similar positive correlation can also be observed in \figref{fig:box_flow_cycle}(c) and \figref{fig:box_flow_cycle}(d), which display the variations in cycle length distributions across distinct traffic flows. 
However, we notice that with a further increase in cycle length, traffic flow tends to decrease. A similar situation is also shown in \figref{fig:box_flow_cycle}(c). This can be attributed to the fact that although a positive correlation is expected between traffic flow and cycle length for the same lane, the lanes with the longest cycle lengths may not necessarily correspond to the highest traffic flows due to different traffic conditions and signal control strategies among these lanes, and vice versa.

\begin{figure*}[tb]
  \centering
  \subfigure[{Spatial distribution of camera sensors installed in lanes entering smart intersections.}]{
    \includegraphics[width=0.66\columnwidth]{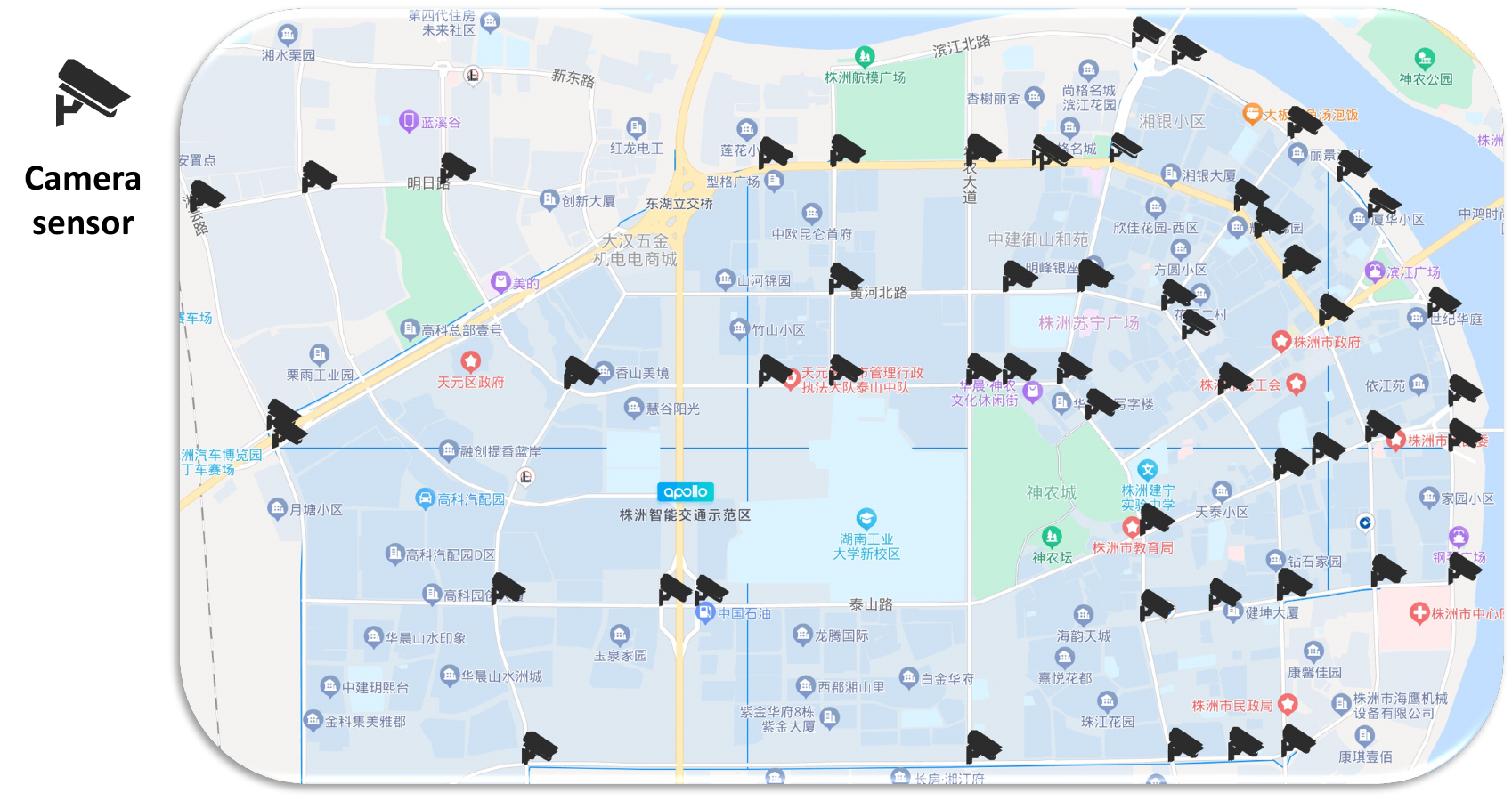}}\hspace{2mm}
  \subfigure[{Spatial distribution of sensors' average traffic flows. Brighter colors represent larger values.}]{
    \includegraphics[width=0.6\columnwidth]{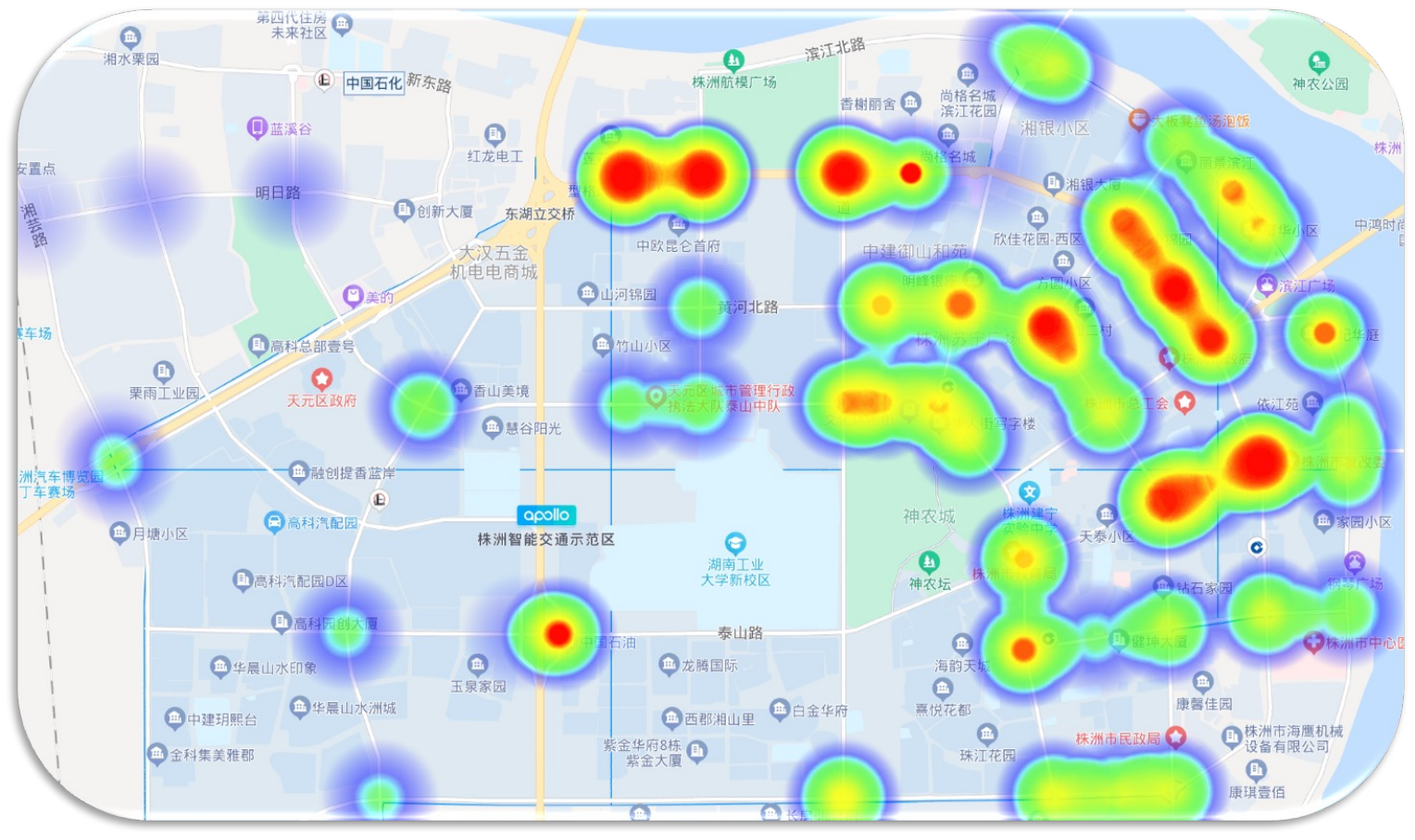}}\hspace{2mm}
  \subfigure[{Spatial distribution of sensors' average cycle lengths. Brighter colors represent larger values.}]{
    \includegraphics[width=0.6\columnwidth]{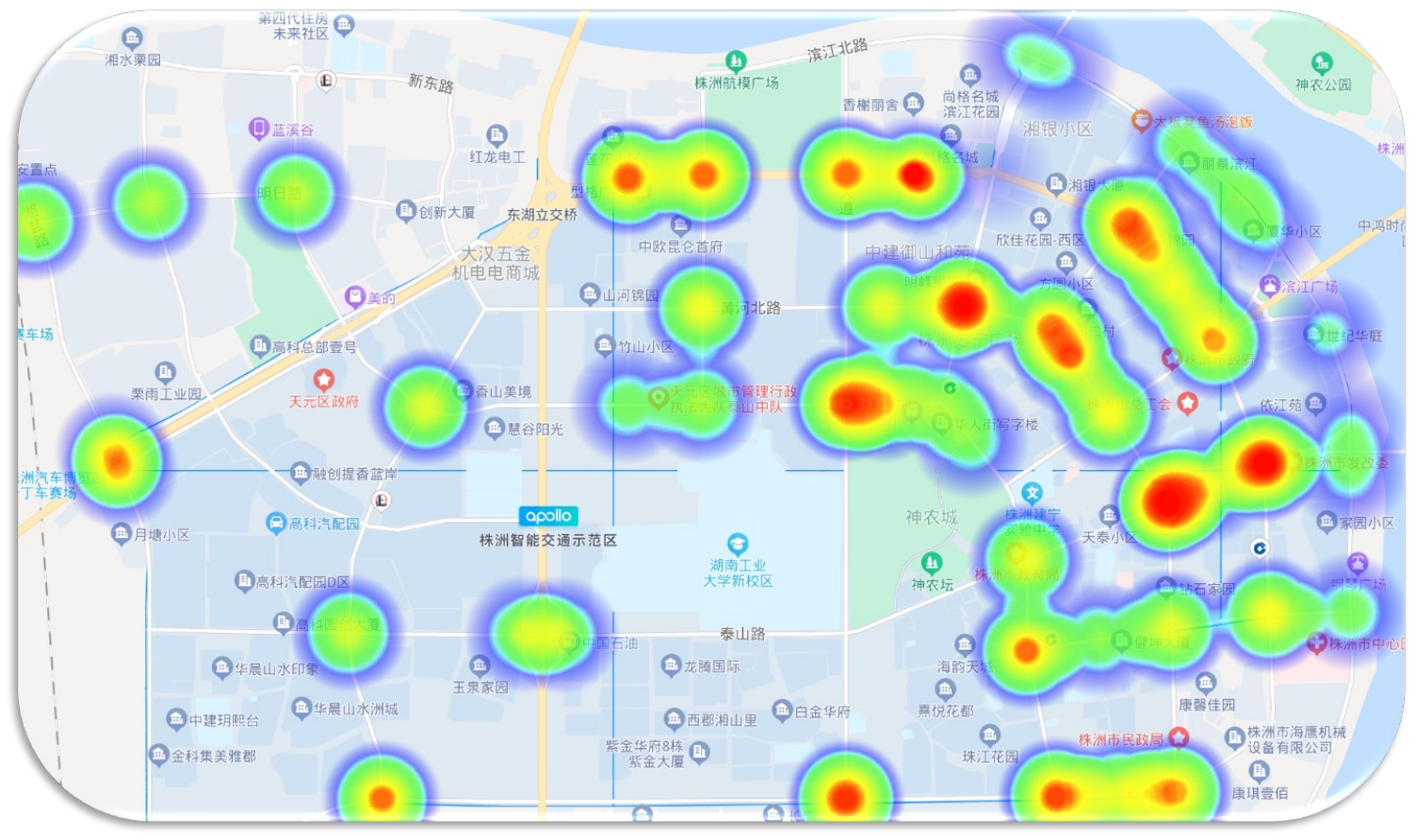}}
  \vspace{-3mm}
  \caption{Spatial distributions of camera sensors and corresponding average traffic flows and cycle lengths on \zhuzhou.} 
  \vspace{-2mm}
  \label{fig:spatial_dist}
\end{figure*}

Additionally, \figref{fig:spatial_dist} displays the spatial distributions of camera sensors and corresponding average traffic flows and cycle lengths on \zhuzhou as a representative. It can be noticed that both traffic flows and cycle lengths exhibit remarked geographical proximity, indicating that neighboring sensors tend to have similar traffic states. This finding provides partial justification for the effectiveness of the spatial dependency modeling component, AGDN, in the irregular traffic forecasting task.

\subsection{\textbf{Baselines}} \label{app:baselines}
We compare our approach with the following twelve baselines. 
\rev{
These baseline models take the same inputs as \model by directly utilizing observed traffic state measurements. 
All these models aim to predict both traffic flow and cycle lengths by optimizing the hybrid loss function in equation \equref{equ:hybridloss}. 
The autoregressive models, \ie~GRU, T-LSTM, GRU-D, and DCRNN, iteratively predict the next step traffic states based on their previous predictions. Since the other non-autoregressive models require the predicted sequence length to be fixed, to enable variable-length sequence prediction, we allow them to predict in a semi-autoregressive way that they iteratively predict a fixed-length sub-sequence based on observed and previously predicted sequences. The prediction step size is set the same as ours to ensure a fair comparison.
} 
We carefully tuned major hyper-parameters of each baseline based on their recommended settings for better performance on our datasets.

\textbf{LAST} predicts future traffic states using the last historical traffic state measurement of each sensor.
\textbf{HA} predicts future traffic states using the average of each sensor's historical traffic state measurements.
\textbf{TCN}~\cite{bai2018empirical} is the temporal convolutional network consisting of causal and dilated convolutions. We apply it to our datasets by padding or intercepting all the sequences to a fixed length. We stack 6 temporal convolution layers with filter size of 3.
\textbf{GRU}~\cite{chung2014empirical} is a powerful variant of recurrent neural networks with a gated recurrent unit. 
\textbf{T-LSTM}~\cite{baytas2017patient} is a time-aware Long-Short Term Memory~(LSTM) model with memory decomposition for irregular time series classification. We modify it to predict traffic states using a LSTM-based decoder.
\textbf{GRU-D}~\cite{che2018recurrent} improves GRU with a time-aware decay mechanism for irregular time series classification. We modify it to predict traffic states using a GRU-based decoder.
\textbf{mTAND}~\cite{shukla2020multi} is a state-of-the-art transformer-based approach for irregularly sampled multivariate time series classification and interpolation tasks. It adopts multi-time attention with time embedding to produce a fixed-length representation of a variable-length time series. The reference point number is set to 64. 
\textbf{Warpformer}~\cite{warpformer2023} is another state-of-the-art transformer-based approach for irregularly sampled multivariate time series classification tasks. It employs a warping module to unify inputs and a doubly self-attention module for representation learning. We set the lengths of warp layers to 0, 24.
\textbf{DCRNN}~\cite{li2018diffusion} is a representative approach based on GNNs and RNNs for classical traffic forecasting tasks, which replaces the matrix multiplications in GRU with a graph convolution operation.
The used graph structure is the same as \model, and the diffusion step is set to 1.
To apply DCRNN to our problem, we pad the input traffic sequences of all nodes to the same length. 
\textbf{GWNet}~\cite{wu2019graph} is a representative approach based on GNNs and CNNs for classical traffic forecasting. It stacks multiple spatial-temporal blocks that are constructed by the graph convolution layer and gated TCN layer, where the graph convolution is performed on the combination of pre-defined and self-learned adjacency matrix. The pre-defined graph structure is the same as \model. We stack 3 blocks with 4 convolution layers and set the convolution filter size to 3. It adopts the same padding strategy as DCRNN.
\textbf{STAEformer}~\cite{liu2023spatio} is a state-of-the-art approach for classical traffic forecasting based on transformer and spatio-temporal adaptive embeddings. We follow the recommended settings the authors give for embedding dimensions and the number of layers and heads and adopts the same padding strategy as DCRNN.
\textbf{PDFormer}~\cite{pdformer2023} is another state-of-the-art transformer-based approach for classical traffic forecasting. It adopts self-attentions for both spatial and temporal dependencies modeling. A graph-masked self-attention mechanism is employed to capture both geographic and semantic spatial dependencies and a delay-aware feature transformation module is used to model the time delay in spatial information propagation.
The depth of encoder layers is set to 2. It adopts the same padding strategy as DCRNN.

\end{document}